%% file: NLPXAI-main.tex
\documentclass[10pt,journal, compsoc]{IEEEtran}
\usepackage[nocompress]{cite} 
\usepackage[bookmarks,backref=true,linkcolor=black]{hyperref} 
\usepackage{verbatim} 
\usepackage{ulem} 
\usepackage{makecell}
\usepackage{mathptmx}
\usepackage{dblfloatfix}
\usepackage[pdftex]{graphicx}
\usepackage{epstopdf}
\usepackage{times}
\usepackage[ruled,vlined]{algorithm2e}  
\usepackage{amssymb,amsmath}
\usepackage{multirow}
\usepackage{subfigure}
\usepackage{paralist}
\usepackage{tikz}
\usepackage{multirow}
\usepackage{color}
\usepackage{setspace}
\usepackage{microtype,hyphenat,balance}
\usepackage{array}
\usepackage{url}
\usepackage{bm}
\usepackage{overpic}
\usepackage{contour}

\usepackage{amsfonts}
\usepackage{bbm}
\usepackage{diagbox}
\usepackage{url}
\usepackage{wrapfig}
\usepackage{microtype,hyphenat,balance,booktabs}

\usepackage{ragged2e}

\def \etal {{\emph{et al}.\thinspace}}

\newcommand{\xitingrevision}[1]{\textcolor{black}{#1}}

\newcommand{\xiting}[1]{\textcolor{black}{#1}}
\newcommand{\zhen}[1]{\textcolor{black}{#1}}
\newcommand{\lizhen}[1]{\textcolor{black}{#1}}
\newcommand{\zhennew}[1]{\textcolor{black}{#1}}

\newcommand{\shixia}[1]{\textcolor{black}{#1}}
\newcommand{\jw}[1]{\textcolor{black}{#1}}
\newcommand{\jwR}[1]{\textcolor{black}{#1}}
\newcommand{\zhengyan}[1]{\textcolor{black}{#1}}
\newcommand{\doc}[1]{\textcolor{black}{#1}}
\newcommand{\liu}[1]{\textcolor{black}{#1}}
\newcommand{\liushixia}[1]{\textcolor{black}{#1}}

\newcommand{\xitingr}[1]{\textcolor{black}{#1}}
\newcommand{\jwrevision}[1]{\textcolor{black}{#1}}
\newcommand{\grammarly}[1]{\textcolor{black}{#1}}

\usepackage[nocompress]{cite} 
\usepackage[backref=true,linkcolor=black]{hyperref} 
\usepackage{caption}
\usepackage{epstopdf}
\usepackage{times}
\usepackage[ruled,vlined]{algorithm2e}  
\usepackage{amssymb,amsmath}
\usepackage{multirow}
\usepackage{subfigure}
\usepackage{paralist}
\usepackage{tikz}
\usepackage{multirow}
\usepackage{color}
\usepackage{setspace}
\usepackage{microtype,hyphenat,balance}
\usepackage{array}
\usepackage{url}
\usepackage{bm}

\usepackage{amsfonts}
\usepackage{bbm}
\usepackage{diagbox}
\usepackage{url}
\usepackage{wrapfig}

\usepackage{ragged2e}

\ifCLASSOPTIONcompsoc
  \usepackage[nocompress]{cite}
\else
  \usepackage{cite}
\fi
\ifCLASSINFOpdf
\else
\fi
\begin{document}
%
\title{A Unified Understanding of Deep NLP Models for Text Classification}
%
%
%
%

\author{Zhen Li, Xiting Wang, Weikai Yang, Jing Wu, Zhengyan Zhang, \\Zhiyuan Liu, Maosong Sun, Hui Zhang, Shixia Liu
\IEEEcompsocitemizethanks{
\IEEEcompsocthanksitem Z.~Li, W.~Yang, Z.~Zhang, Z.~Liu, M.~Sun, H.~Zhang, and S.~Liu are with Tsinghua University. H.~Zhang is the corresponding author.
\IEEEcompsocthanksitem W.~Yang is with Microsoft Research Asia. 
\IEEEcompsocthanksitem J.~Wu is with Cardiff University.
}
}

\input{0-abstract.tex}


\maketitle
\IEEEdisplaynontitleabstractindextext
\IEEEpeerreviewmaketitle

{
\fontsize{10}{10} 
\input{1-introduction.tex}
\input{2-related.tex}
\input{3-requirement.tex}

\input{4-visualization.tex}
\input{5-application.tex}

\input{6-discussion.tex}
\input{7-conclusion.tex}

}

\ifCLASSOPTIONcompsoc
 \section*{Acknowledgments}
\else
 \section*{Acknowledgment}
\fi
This work was supported by the National Key R\&D Program of China under Grant 2020YFB2104100, the National Natural Science Foundation of China under grants U21A20469, 61936002, grants from the Institute Guo Qiang, THUIBCS, and BLBCI, and in part by Tsinghua-Kuaishou Institute of Future Media Data.


\ifCLASSOPTIONcaptionsoff
  \newpage
\fi

\bibliographystyle{IEEEtran}
\bibliography{reference}
\vspace{-12mm}

\begin{IEEEbiography}
[{\includegraphics[width=1in,height=1.25in,clip,keepaspectratio]{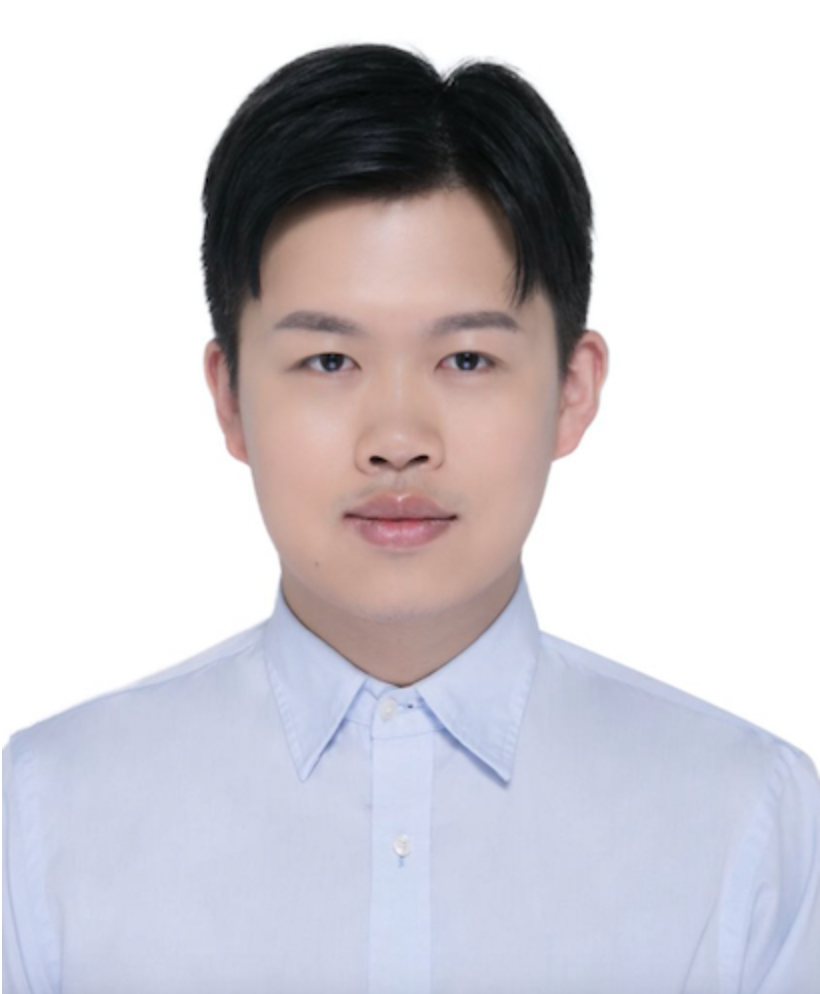}}]
{{Zhen Li}} is a first-year Ph.D. student of Software School, Tsinghua University. His research interest is explainable artificial intelligence. He received a B.S. degree from
Tsinghua University and a M.Phil. degree from Hong Kong University of Science and Technology.
\end{IEEEbiography}

\begin{IEEEbiography}
[{\includegraphics[width=1in,height=1.25in,clip,keepaspectratio]{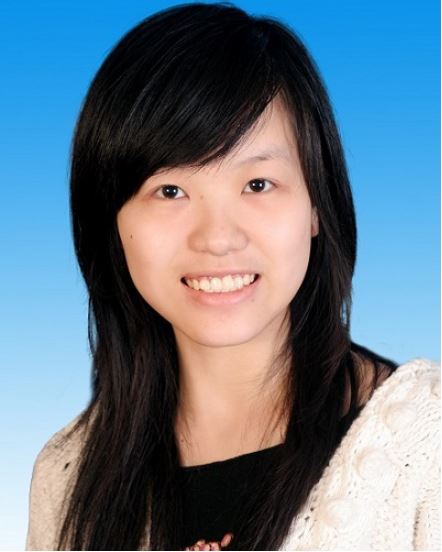}}]{{Xiting Wang}} is a senior researcher at Microsoft Research Asia. Her research interests include explainable machine learning and visual text analytics. She has published academic papers on reputable international conferences and journals in her research area, such as KDD, TKDE, AAAI, IJCAI, TVCG and VAST. One of her first author papers has been chosen as the TVCG spotlight article for Dec. 2016. She is a senior program committee member of AAAI and is a program committee member of many top conferences.\looseness=-1
\end{IEEEbiography}

\begin{IEEEbiography}
[{\includegraphics[width=1in,height=1.25in,clip,keepaspectratio]{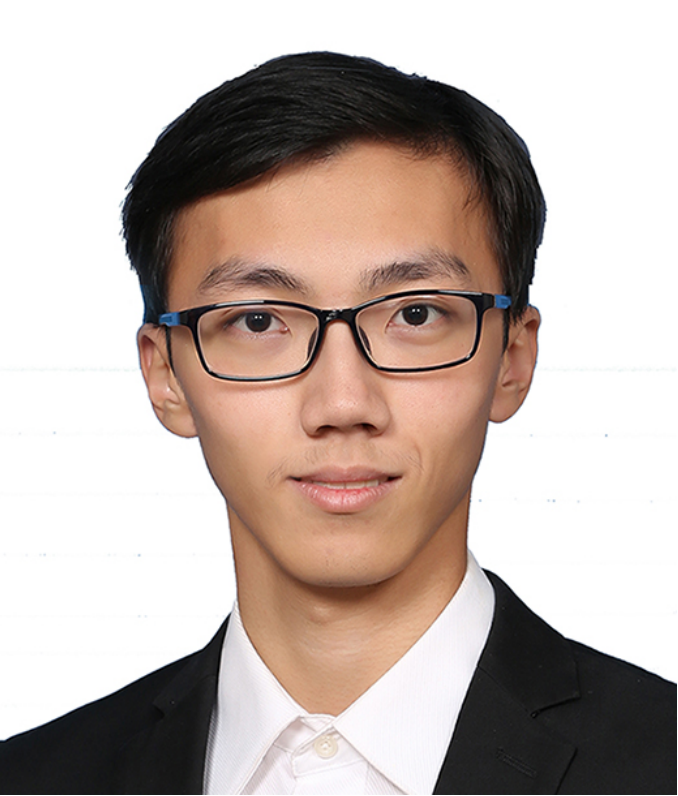}}]
{{Weikai Yang}} is a third-year Ph.D. student at Tsinghua University. His research interests lie in integrating the Machine Learning into Visual Analytics, which can facilitate the understanding of large-scale data and make it easier for the practitioners to use the machine learning techniques.
\end{IEEEbiography}

\begin{IEEEbiography}
[{\includegraphics[width=1in,height=1.25in,clip,keepaspectratio]{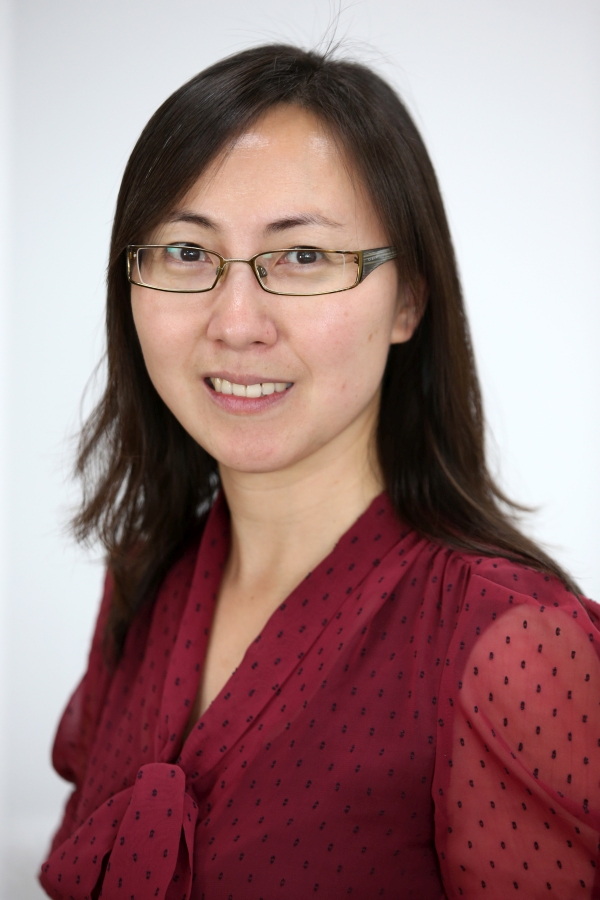}}]{Jing Wu} is a lecturer in computer science and informatics at Cardiff University, UK. Her research interests are in computer vision and graphics including image-based 3D reconstruction, face recognition, machine learning and visual analytics. She received BSc and MSc from Nanjing University, and Ph.D. from the University of York, UK. She serves as a PC member in CGVC, BMVC, etc.
\end{IEEEbiography}

\begin{IEEEbiography}[{\includegraphics[width=1in, height=1.25in, clip, keepaspectratio]{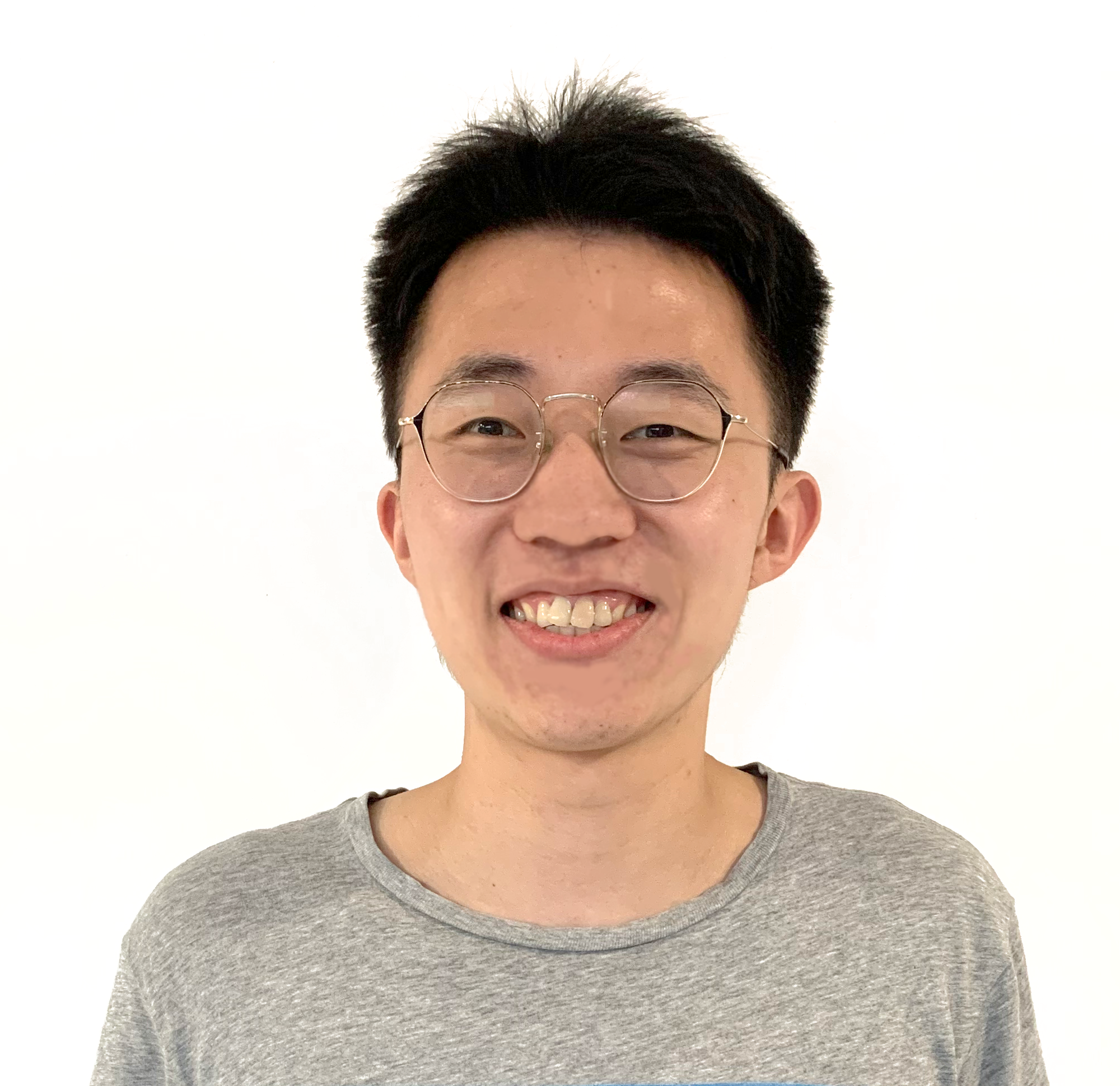}}]{Zhengyan Zhang}
is a second-year Ph.D. student of the Department of Computer Science and Technology, Tsinghua University. His research interests include natural language processing and social computing. He has published papers in international conferences and journals, including ACL, EMNLP, and TKDE.
\end{IEEEbiography}

\begin{IEEEbiography}[{\includegraphics[width=1in, height=1.25in, clip, keepaspectratio]{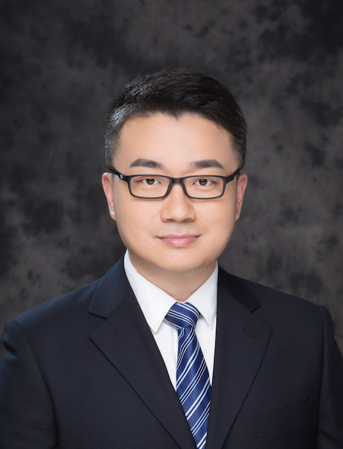}}]{Zhiyuan Liu}
is an associate professor at Tsinghua University. He got his BEng degreeand his Ph.D. from Tsinghua University. His research interests are natural language processing, information extraction, knowledge graphs, and social computation. He has published over 80 papers in international journals and conferences, including ACM/IEEE Transactions, AAAI, IJCAI, ACL, and EMNLP. He has also served as PC/Area Chair of several international conferences, including ACL, EMNLP, WWW, CIKM, COLING, etc.
\end{IEEEbiography}

\begin{IEEEbiography}[{\includegraphics[width=1in, height=1.25in, clip, keepaspectratio]{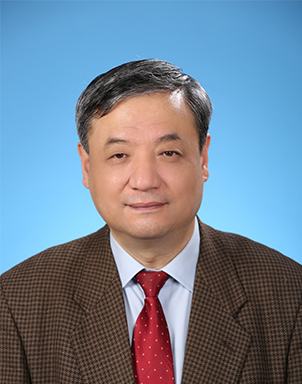}}]{Maosong Sun}
is a professor at Tsinghua University. He got his BEng degree and MEng degree from Tsinghua University, and got his Ph.D. degree from City University of Hong Kong. His research interests include natural language processing, Chinese computing, Web intelligence, and computational social sciences. He serves as a vice president of the Chinese Information Processing Society, the council member of China Computer Federation, and the Editor-in-Chief of the Journal of Chinese Information Processing.
\end{IEEEbiography}

\begin{IEEEbiography}
[{\includegraphics[width=1in,height=1.25in,clip,keepaspectratio]{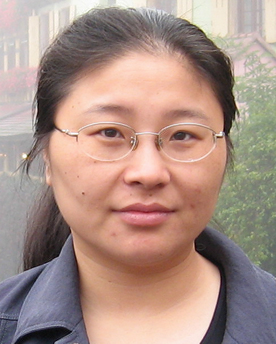}}]{Hui Zhang}
is an Associate Professor at School of Software, Tsinghua University, China. She received her B.Sc. and Ph.D. in Computer Science from Tsinghua University, in 1997 and 2003, respectively. Her research interests include computer aided design and computer graphics.
\end{IEEEbiography}

\begin{IEEEbiography}
[{\includegraphics[width=1in,height=1.25in,clip,keepaspectratio]{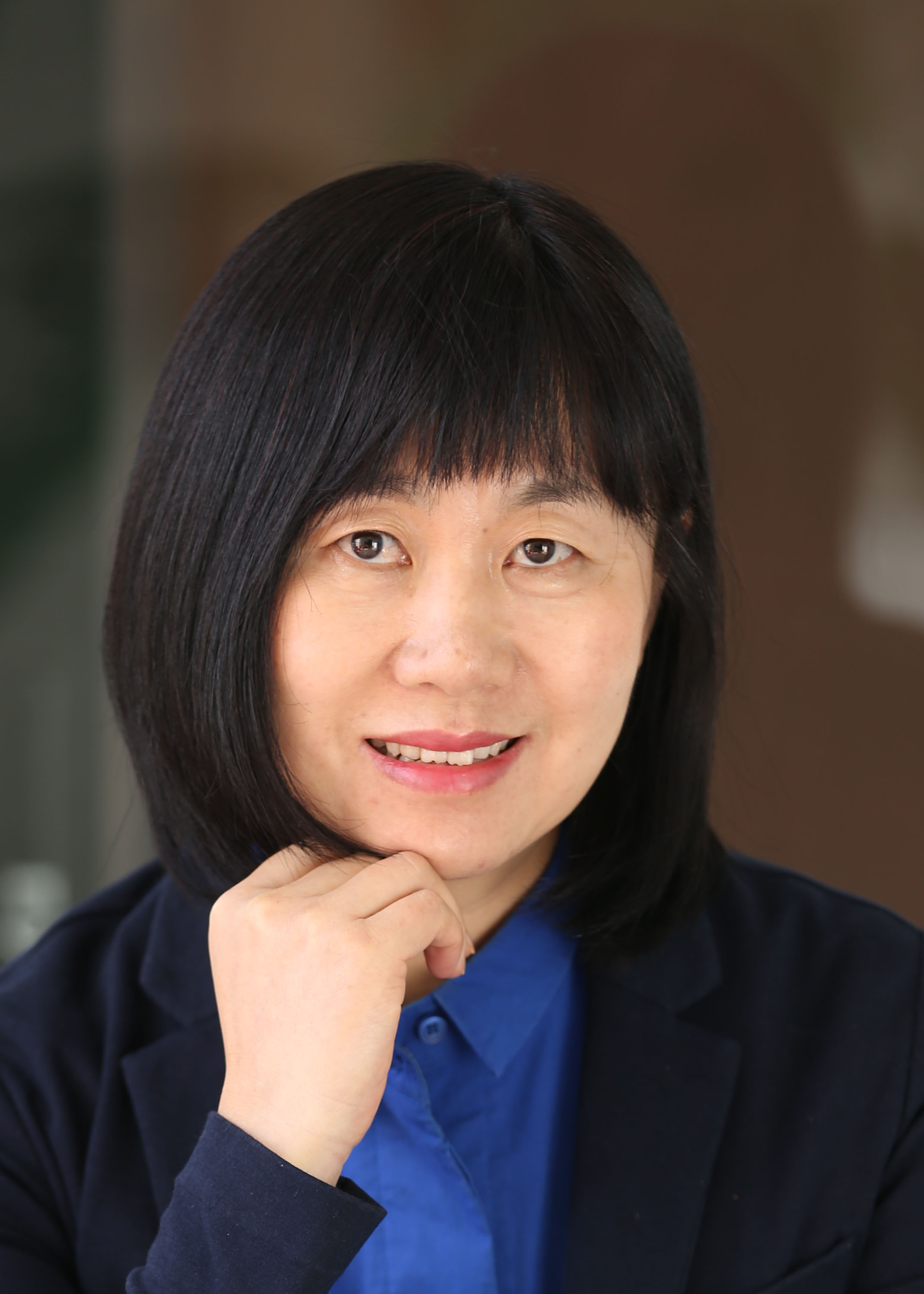}}]{Shixia Liu}
is a professor at Tsinghua University. Her research interests include explainable artificial intelligence, visual text analytics, and text mining. She worked as a research staff member at IBM China Research Lab and a lead researcher at Microsoft Research Asia.
She received a B.S. and M.S. from Harbin Institute of Technology, a Ph.D. from Tsinghua University.
She is a fellow of IEEE and an associate editor-in-chief of IEEE Trans. Vis. Comput. Graph.
\end{IEEEbiography}

\end{document}


\maketitle
\vspace{-22mm}


\subsection*{Part One: Current Practice}
\Qitem{ \Qq{How many years of experience do you have in deep neural networks for NLP? 
[Select one of the following] }
\begin{Qlist}
\item Less than 1 year
\item 1 to 3 years
\item 3 to 5 years
\item More than 5 years
\end{Qlist}
}
\Qitem{ \Qq{Which types of deep neural networks do you 
use? 
[Select all that apply] }
\begin{Qlist}
\item Convolutional Neural Networks
\item Recurrent Neural Networks and Long Short-Term Memory Networks
\item Transformers (Train from scratch)
\item Pre-trained models (e.g., BERT, RoBERTa, GPT)
\item Other: \Qline{4cm}
\end{Qlist}
}

\Qitem{ \Qq{Which tasks do you use deep neural networks for? 
[Select all that apply] }
\begin{Qlist}
\item Text classification
\item Text generation
\item Text summary
\item Machine question and answer
\item Machine translation
\item Other: \Qline{4cm}
\end{Qlist}
}

\Qitem{ \Qq{How many types of deep neural networks (e.g., LSTM, CNN, Transformer, BERT) have been used in your recent projects? } 

\vspace{2mm}
\Qline{16cm}}

\Qitem{ \Qq{How many models with different hyperparameters (e.g., number of layers, latent vector size) have been used in your recent projects? } 

\vspace{2mm}
\Qline{16cm}}


\Qitem{ \Qq{Which problems do you 
encounter when training deep NLP models? 
[Select all that apply] }
\begin{Qlist}
\item Overfitting
\item Underfitting
\item Disappearing gradients or exploding gradients
\item Training data quality problems (e.g., uneven data distribution, incorrect labels)
\item Misunderstanding of specific samples
\item Misunderstanding of specific words
\item Other: \Qline{4cm}
\end{Qlist}
}

\Qitem{ \Qq{How do you analyze the results and identify the problems of a model? 
[Select all that apply] }
\begin{Qlist}
\item By analyzing the accuracy of the model
\item By analyzing the trend of loss during model training
\item By analyzing the training logs of the model, including the intermediate results and gradients
\item By using visualization tools (e.g., Tensorboard, BertViz)
\item Other: \Qline{4cm}
\end{Qlist}
}


\Qitem[]{ \Qq{ Are you interested in a tool that can help you analyze 
different models in a unified manner?} 

Not 
interested / Slightly interested / Moderately interested / Interested / Extremely interested

{Not 
interested \Qrating{5} Extremely interested}}

\subsection*{Part Two: Key Features Needed}


\Qitem[]{ \Qq{Do you think the following features can help you understand the model and further improve the performance? Please rate the importance. 
}

Unimportant / Slightly important / Moderately important / Important / Very important
}

\Qitem[a]{ \Qq{Analyzing the overall model behavior on the entire training set.}

{Unimportant \Qrating{5} Very important}}
\Qitem[b]{ \Qq{Identifying the key samples for understanding and debugging the model.}

{Unimportant \Qrating{5} Very important}}
\Qitem[c]{ \Qq{Understanding the prediction of a given sample.}

{Unimportant \Qrating{5} Very important}}
\Qitem[d]{ \Qq{Identifying the key words for understanding and debugging the model.}

{Unimportant \Qrating{5} Very important}}
\Qitem[e]{ \Qq{Understanding the impact of a given word on the model prediction.}

{Unimportant \Qrating{5} Very important}}
\Qitem[f]{ \Qq{Understanding the model behavior at the layer level.}

{Unimportant \Qrating{5} Very important}}
\Qitem[g]{ \Qq{Understanding the model behavior at the neuron level.}

{Unimportant \Qrating{5} Very important}}
\Qitem[h]{ \Qq{Understanding the impact of other model-specific designs, such as specific activation functions, recurrent cells, convolution layers.} 

{Unimportant \Qrating{5} Very important}}

\Qitem[i]{ \Qq{Please provide other important features 
(if any).} 

\vspace{1em}
\Qline{16cm}}
\clearpage

\Qitem[]{ \Qq{Do you think the following analysis processes are 
effective for 
understanding model behaviors? Please rate the effectiveness.}

Completely ineffective / Somewhat ineffective / Somewhat effective / Effective / Very effective
}

\Qitem[a]{ \Qq{Corpus-level analysis: 1) understand what information the model tends to discard or 
retain in the overall sample set; 2) find unexpected behaviors (e.g., discarding important information or retaining irrelevant information); 3) refine the model accordingly. 
}

{Completely ineffective \Qrating{5} Very effective}}
\Qitem[b]{ \Qq{Sample-level analysis: 1) find 
mispredicted samples; 2) select other samples with similar semantics and analyze the root causes for misprediction; 3) refine the model accordingly. 
}

{Completely ineffective \Qrating{5} Very effective}}
\Qitem[c]{ \Qq{Word-level analysis: 1) find the words wrongly understood by the model; 2) select the samples containing these words and analyze the root causes for misunderstanding; 3) refine the model accordingly. 
}

{Completely ineffective \Qrating{5} Very effective}}
\Qitem[d]{ \Qq{Decision-boundary-based analysis: 1) find 
mispredicted samples near the decision boundary; 
2) analyze the root causes for misprediction; 3) refine the model accordingly.
}

{Completely ineffective \Qrating{5} Very effective}}

\Qitem[e]{ \Qq{Please provide other effective analysis processes (if any).}

\vspace{2mm}
\Qline{16cm}

\Qline{16cm}
\vspace{2mm}
}


\maketitle


We experiment with seven datasets and find the best 
value for $\xi$.
According to the experts, a good value of $\xi$ should well differentiate class-relevant and class-irrelevant words.
For example, words like ``ipo" and ``spending" should be considered relevant with ``business", words like ``search" and ``gates" should be considered relevant with ``sci/tech", and words like ``time" and ``can" should be class-irrelevant.
Based on this criteria, the experts find the best range for $\xi$ together with the experts.
As shown in Table~\ref{table:parameter-experiment}, the best value for $\xi$ usually increases with increasing text length.
For datasets with shorter text (average number of characters smaller than 300), the range 0.01 to 0.03 is usually a good choice. For datasets with longer text (average number of characters larger than 300), the range 0.03 to 0.05 is a good choice. 
We suspect that this is because for longer text, the information is scattered across more words, resulting in a larger variance of mutual information and thus the requirement for 
a larger margin $\xi$. 

\begin{table}[htbp]
\centering
\caption{The best 
range for parameter $\xi$ on seven different datasets}
\begin{tabular}{llll}
\hline
Dataset & Number of chars & Number of tokens & Best $\xi$ \\ \hline
SST-2~\cite{socher2013recursive} & 53.5 & 13.3 & 0.01-0.02 \\
AG News~\cite{zhang2015characterlevel} & 236.5 & 53.3 & 0.01-0.03 \\
BBC News~\cite{greene06icml} & 264 & 56.98 & 0.02-0.03 \\
BBC Sports News~\cite{greene06icml} & 295.8 & 62.8 & 0.02-0.03 \\
Amazon Product Data~\cite{he2016ups} & 703 & 159.6 & 0.04-0.05 \\
IMDB Reviews~\cite{maas2011acl} & 907 & 216.3 & 0.04-0.05 \\
Medical~\cite{medical-text} & 1233 & 281.3 & 0.04-0.05 \\ \hline
\end{tabular}
\label{table:parameter-experiment} 
\end{table}

\bibliographystyle{IEEEtran}
\bibliography{appendix-ref}

%% file: 0-abstract.tex
\IEEEtitleabstractindextext{%
\justify
\begin{abstract}
The rapid development of deep natural language processing (NLP) models for text classification has led to an urgent need for a unified understanding of these models proposed individually. 
Existing methods cannot meet the need for understanding different models in one framework due to the lack of a unified measure for explaining both low-level (e.g., words) and high-level (e.g., phrases) features. 
We \doc{have developed} a visual analysis tool, DeepNLPVis, to enable a unified understanding of NLP models for text classification. 
The key idea \jwR{is a mutual information-based measure,} 
which provides quantitative explanations on how each layer of a model maintains the information of input words in a sample.  
We model the intra- and inter-word information at each layer \jwR{measuring} the importance of a word to the final prediction as well as the relationships between words, such as \jwR{the formation of} phrases.
A multi-level visualization, \jwR{which consists} 
\doc{of} a corpus-level, \jwR{a} sample-level, and \jwR{a} word-level visualization, supports the analysis from the overall training set to individual samples.
Two case studies on classification tasks and comparison between models \liushixia{demonstrate that DeepNLPVis can help users effectively identify potential problems caused by samples and model architectures and then make informed improvements.}
\end{abstract}

\begin{IEEEkeywords}
Explainable AI, visual debugging, visual analytics, deep NLP model, information-based interpretation
\end{IEEEkeywords}}

%% file: 1-introduction.tex
\section{Introduction}
\maketitle
Text classification is a fundamental task in natural language processing (NLP) and has been under rapid development to assist our everyday communication~\cite{minaee2021deep}. 
In recent year\lizhen{s}, different deep NLP models from CNN-based~\cite{kim2014convolutional}, LSTM-based~\cite{strobelt2017lstmvis}, and Transformer/attention-based~\cite{vaswani2017attention} have been consecutively proposed  \jwrevision{to improve} 
the performance of text classification tasks.
However, along with the improved performance is the increasing complexity of the model architecture, which poses difficulties for model developers not only in training the \liu{model}, but also in debugging when the performance is not as expected.

For example, the recent BERT \jwR{model}~\cite{devlin2018bert} contains hundreds of millions of parameters. Training such a \jwR{model} 
from scratch requires massive data and computing resources that are unaffordable to most NLP \jwR{developers.} 
A training schema with pre-training and fine-tuning is thus getting popular.
Starting from a \xitingrevision{pre-trained} model (e.g., a pre-trained BERT), model developers fine-tune the \jwR{model} 
to the \jwR{end classification} task with specific input-output designs. 
They usually follow two approaches to improve the model performance.
One is to augment the training data and improve the data quality with the model architecture unchanged. 
In machine learning, it has been proposed that ``$80\%$ data + $20\%$ model = better machine learning"\cite{kim2021playing}, which demonstrates the importance of data quality. 
Label errors, missing samples, and sample bias are all factors that affect classification performance~\cite{yuan2021survey}.
The other approach is to make slight changes to the model architecture. 
For example, inserting ``adapter layers''  \liushixia{between specific layers of the model} has been shown effective 
\grammarly{in improving} model performance~\cite{wang2020k}.

Effective improvement from either of the above approaches will require the model developers to understand the model's working mechanism, and in turn to identify the deficiencies for informed augmentation of the data and/or adaptation of the model architecture. 
Several visualization tools have been developed to facilitate the understanding of a specific deep NLP model, such as RNNVis~\cite{ming2017understanding} and Attention Flows~\cite{derose2021attention} for RNN-based and attention-based NLP models, respectively. 
Our work follows this direction to assist in understanding deep NLP \liu{models for text classification}. 
However, we argue that a tool for a specific \liu{model} may restrict the model developer's choice of the most suitable \liu{model for \jwR{a specific task}}. 
Currently, there is a trend to revisit deep learning models~\cite{tolstikhin2021mlp}, and it has been found some simple models can actually achieve competing performance as more complex models. 
Given the high computing demand of pre-trained \liu{models}, it would be interesting to know whether simpler \liu{models}, such as LSTM or CNN-based \liu{models}, are potential alternatives at least for \liu{some scenarios \jwR{in classification tasks}}. 
\jwrevision{For example, improving the accuracy of these simpler models will make them deployable on portable devices with limited computing resources, such as mobile phones.} 
It \jwrevision{thus} naturally arouses the interest to revisit different \liu{classification models} in NLP. 
However, due to the diverse model architectures, this is non-trivial.
A fundamental requirement is to unify the understanding of these \liu{models}' working mechanisms.
With this in mind, although our work follows the direction of developing visualization tools for deep NLP \liu{models}, it has a unique focus on a unified understanding across \liu{different classification models}. 
It aims to help model developers \lizhen{better understand} the strengths and weaknesses of different NLP \liu{models} and make informed improvements.

We thus develop DeepNLPVis, an interactive visual analysis tool to help model developers gain a unified understanding of different \jwR{NLP models for text classification,}  
quickly identify problems, and make informed improvements. 
We propose to improve the mutual information-based measure in~\cite{guan2019towards} to explain the information learned by intermediate layers, 
including \jwR{both intra-word information (e.g., word contribution to classes) and inter-word information (e.g., relationships between words).}
With this measure, DeepNLPVis adopts a coordinated multiple-level visualization connecting the analysis from the overall training corpus to individual samples and words (Fig.~\ref{fig:teaser}). 
\jwrevision{At the corpus-level, a class view shows the overall model performance on all} \zhennew{classes}\jwrevision{, from which the user can select two classes to explore the corpus against predictions on the two classes in the distribution view. 
The sample-level visualization supports analyzing a sample in terms of both intra-word and inter-word information. 
And the word-level visualization supports examining words in terms of word contribution and word meaning. }
The interactive analysis enabled by the coordinated visualization helps users explore \liu{model} deficiencies and identify the root cause of low performance. The demo is available at \href{http://nlpvis.thuvis.org/}{\textcolor{blue}{http://nlpvis.thuvis.org/}}.

In summary, the main contributions of this work are:
\begin{compactitem}
\item\noindent{A mutual-information-based visual analysis tool for efficient identification and diagnosis of problems in deep NLP \liu{models} for text classification.}
\item\noindent{An information-based sample interpretation method for simultaneously understanding \jwR{the intra-word and inter-word information in a unified way.}
}
\item\noindent{A three-level visualization consisting of a corpus-level visualization for quickly identifying samples and words of interest, and a sample-level and a word-level visualization to disclose \jwR{the intra-word and inter-word information and their changes across layers.} 
}
\end{compactitem}

%% file: 2-related.tex
\section{Related Work}


\subsection{Machine Learning for Understanding NLP Models}
Existing machine learning methods for understanding deep NLP models can be categorized into three classes: built-in interpretability methods, post-hoc model-specific methods, and post-hoc model-agnostic methods~\cite{dovsilovic2018explainable,ming2019interpretable}. 
The first category attempts to design self-explanatory \doc{models}.
The second category analyzes some specific \lizhen{architectures} in the model, such as the hidden states or attention heads.
The third category interprets the model by considering model inputs, intermediate layers, and outputs. 
These methods are model-agnostic as they do not make assumptions about the specific model architecture.
Our work is relevant to the third category.
Here we briefly review the works along this line.

Many post-hoc model-agnostic methods utilize an easy-to-interpret model, such as decision trees, to approximate the original model and explain its behavior~\cite{baehrens2010explain}.
Local Interpretable Model-agnostic Explanations~\cite{ribeiro2016should} is \doc{a representative work,} 
which trains a sparse linear model and uses it to explain the black-box model locally.
Recently, Guan~\etal~\cite{guan2019towards} proposed a measure to quantify the information stored in each word.
This measure provides quantitative explanations on \liushixia{the contribution of a word to the final prediction layer by layer.}
Compared with existing methods, the explanations provided by this method are consistent across different NLP models.
Our method extends this unified measure to help understand \liushixia{
which class the word contributes to and the relationships between words (e.g., phrases)}.
\liushixia{We also leverage interactive visualization to visually explain the aforementioned word-related information across layers. 
This helps users quickly identify unusual words/phrases that lead to low performance.}  

\subsection{\xiting{Visualization for} Understanding NLP Models}
\liushixia{Existing visualization methods for visually understanding machine learning models can be categorized into two classes: domain irrelevant~\cite{ren2016squares,zhang2019manifold} and domain specific~\cite{chawla2020,dong2020interactive,Liu2018Analyzing,ji2021usevis,Gehrmann2020Visual,liu2019nlize}. 
Our work is in the second category with a focus in the NLP domain, 
which enables an in-depth understanding of the working mechanism of the model training process.
Thus, we briefly review the works along this line.}  

Earlier efforts focus on utilizing simple and static diagrams, such as heatmaps~\cite{karpathy2016visualizing,li2015visualizing}, to demonstrate which input words play important roles in model prediction.
Later efforts \doc{employ} interactive visualization to analyze the intermediate layers of the models, such as hidden states and attention mechanisms~\cite{hohman2018visual,yuan2021survey}.
Ming~\etal~\cite{ming2017understanding} introduced a visual analysis method for interpreting hidden states of RNNs based on their expected response to the input.
Strobelt~\etal~\cite{strobelt2017lstmvis} developed LSTMVis to explain the hidden state changes and identify similar examples.
RNNbow~\cite{cashman2018rnnbow} visualizes the gradient flow in the training process of RNNs to investigate the learning behavior of a model.
Seq2Seq-Vis~\cite{strobelt2018s} visually analyzes the five black-box stages of the machine translation process.
The attention mechanism is widely used in visual model explanation due to its high interpretability.
Accordingly, RetainVis~\cite{kwon2018retainvis} leverages the attention mechanism to explain how predictions are made and supports the modification of the input or the model to conduct what-if analysis.
Tenney~\etal~\cite{tenney2020language} developed a language interpretability tool, which utilizes the attention mechanism and salience maps to provide a comprehensive understanding of \doc{model behavior}.
More recently, Derose~\etal~\cite{derose2021attention} developed Attention Flows to trace and compare the attention heads in a BERT model.
The system supports both single model analysis and comparison between pre-trained and fine-tuned models.

The aforementioned works have achieved considerable success in understanding deep NLP models. \doc{However,} they mainly focus on a specific type of NLP models (e.g., RNNs or attention-based models).
In comparison, we \doc{have developed} a unified method for understanding different types of NLP \liu{models for text classification}.
Our method enables a deeper understanding of the capabilities of different classification models by revealing the word contribution and word relationships at different layers.
A multi-level visualization is also carefully designed to help users quickly \doc{identify} the \xiting{samples and words of interest} in the context of the overall data distribution and model prediction.

%% file: 3-requirement.tex

\section{Requirement Analysis}\label{Sec:Design}

\subsection{Survey on Practices of Building Deep NLP Models}

We conducted a questionnaire to better understand the current practices of NLP model developers and the key functions they need for efficiently developing deep models for text classification.

\noindent \textbf{\normalsize Designing the questionnaire.}
The questionnaire was designed and iteratively refined with \zhen{five} model developers,
who have varying experience (from one year to \zhen{eight} years) in NLP.
We conducted a 45-60 minute semi-structured interview with each of them.
In the interview, we first asked the model developers to introduce his/her current practices and difficulties. 
Then, we explored the functions that s/he needed for better understanding and debugging \liu{a model}.
Their feedback \doc{was} summarized to create the questionnaire.

\begin{figure}[t]
  \centering
  \includegraphics[width = 1\linewidth]{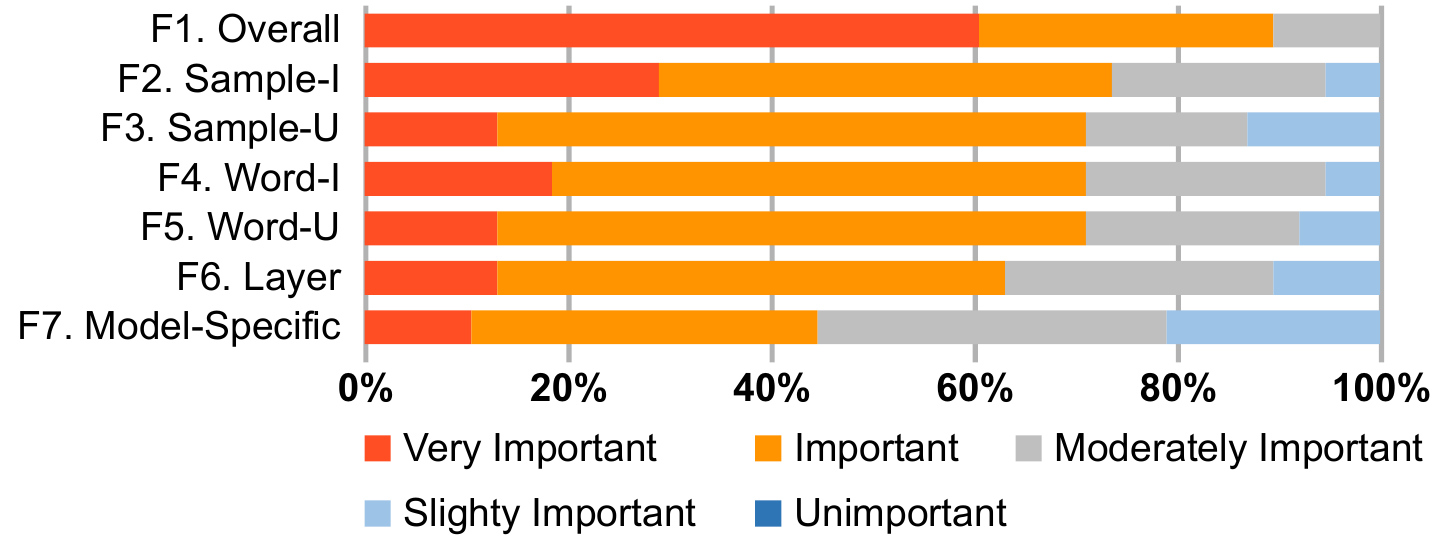}
    \caption{Importance of different functions. Sample-I (or Word-I) refers to the identification of samples (or words) of interest, and Sample-U (or Word-U) denotes understanding a sample (or a word). 
      }
    \label{fig:questionnaire-functions} 
\end{figure}

\noindent \textbf{\normalsize Conducting the questionnaire.}
The questionnaire was distributed to 1) students in three NLP groups of \xiting{a top university} and 2) NLP model developers in \zhen{a major technology company}. 
\liushixia{Out of the 46 returned questionnaires, eight (17.4\%) were discarded due to incomplete responses}. 
Among the participants, 21.1\% had less than 1-year experience in NLP,
44.7\% had 1-3-year experience, 28.9\% had 3-5-year experience, and 5.3\% had 5-10-year experience.
The NLP models they used include Pre-trained Model such as BERT (81.6\%), Transformer (73.7\%), RNN (52.6\%), and CNN (36.8\%).

\underline{\emph{\normalsize Current practices and difficulties}}.
A majority of the participants understood and debugged NLP models by investigating the training loss (89.5\%), logging intermediate results (84.2\%), and observing the changes 
\lizhen{in} model accuracy (76.3\%).
Only 34.2\% of them leveraged tools that \doc{were} 
specifically designed for model understanding, e.g., Tensorboard or BertViz.
They commented that it was difficult to use current tools \jwrevision{for identifying} 
problems in the training data, e.g., incorrect labels and uneven data distribution (78.9\%).
It was also difficult to understand why \doc{a model could not} correctly predict \jwrevision{the labels for certain samples} 
(50.0\%) and why the model incorrectly \doc{understood} \jwrevision{certain words} 
(28.9\%).
\shixia{In addition to} the difficulty in understanding and debugging a single model, the participants also \jw{expressed} the needs \doc{to} effectively \doc{compare} different models.
On average, they experimented with five models in the most recent project.
Many participants have shown interest in a tool that can help investigate NLP models in a unified way and compare the models effectively.
We asked them to rate their \doc{interest} 
according to a 1-5 Likert scale, and 84.2\% \zhen{of} participants returned a rating of 4 (\xiting{interested}) or 5 (extremely interested).

\underline{\emph{\normalsize Key functions needed}}.
We then summarized the key functions that help effectively develop NLP models and asked the participants to rate how important each function is based on a 1-5 Likert scale (\xiting{1: unimportant, 5: very important}). 
Fig.~\ref{fig:questionnaire-functions} shows the rating distribution of different functions.
More than 80\% \doc{of} participants considered examining the overall performance and model behavior on the training data (F1) as (very) important.
The participants were also eager to understand how samples and words contribute to the model performance.
Specifically, more than 70\% \doc{of} participants agreed that it was (very) important to identify key samples (F2) and words (F4) for understanding and debugging a model.  
Over 70\% \zhen{of} participants also \jw{expressed} 
the need \doc{to} 
deeply \doc{understand} 
how each sample (F3) or word (F5) affects the model prediction.
Among the information about \lizhen{model architectures} (e.g., layers, neurons, or recurrent cells), analyzing how \doc{a} model behaves across layers (F6) \doc{attracted} 
the most attention.
\xiting{63.2\%} \zhen{of} participants considered understanding layer-wise evolution to be (very) important.
In comparison, investigating neurons or model-specific \lizhen{architecture} (F7) (e.g., specific activation function, recurrent cell, convolution layer) was not \doc{frequently cited by} the participants.

\begin{figure*}[t]
\centering
 \includegraphics[width=0.9\textwidth]{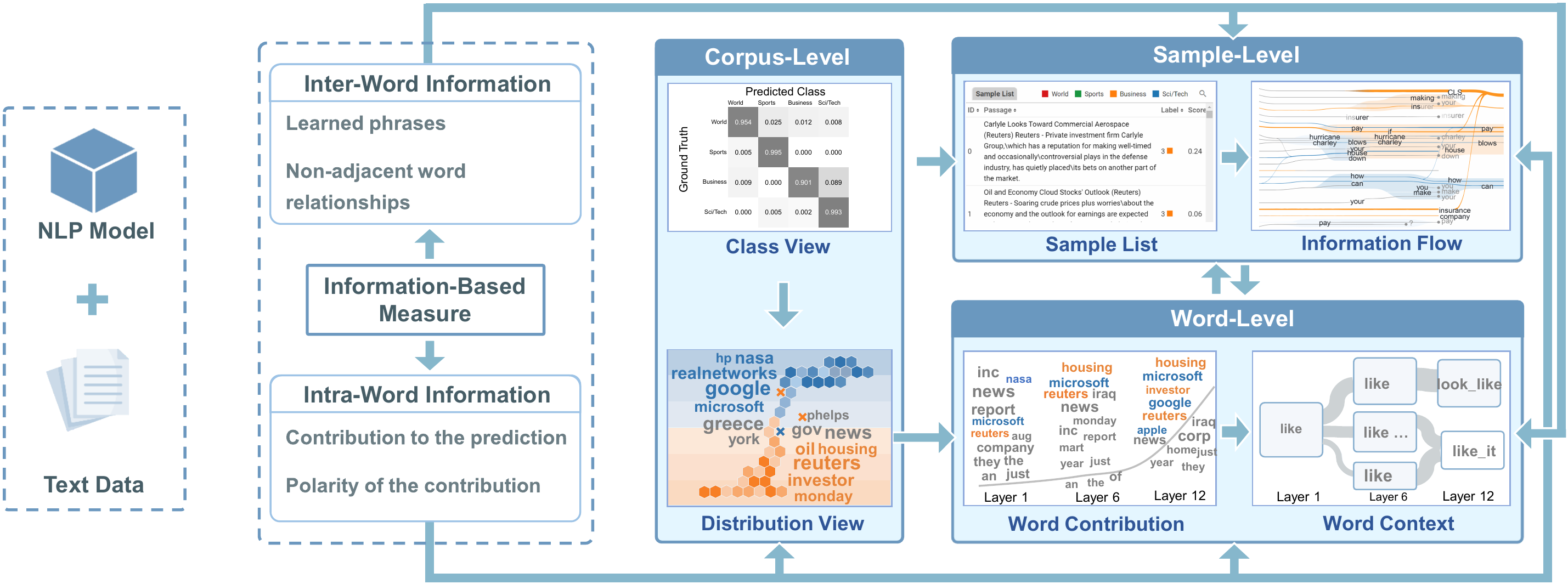}
\caption{The analysis workflow supported by three coordinated visualizations at different levels.}
\label{fig:workflow}
\end{figure*}

\subsection{Design Requirements}
\liushixia{We further} conducted interviews with five \liushixia{experts ($E_1$-$E_5$) selected from \jwrevision{the} 38 questionnaire participants.}
\xitingr{The experts are selected to ensure that they have different levels of expertise and work on various NLP models. 
\xitingr{In particular,} \liushixia{$E_1$} has 1-year experience in NLP, $E_2$ and $E_3$ have 3-year experience, \liushixia{$E_4$} has 5-year experience, and \liushixia{$E_5$} has 10-year experience. 
All experts are familiar with BERT and Transformer. 
\liushixia{$E_4$} and \liushixia{$E_5$} also have experience in training LSTM and CNN.}
Based on the questionnaire survey and interviews, we distilled three-level requirements: corpus, sample, and word.

The \textbf{\normalsize corpus}-level requirement aims to help users obtain a quick overview of the model behavior (F1).

\emph{\normalsize R1. Exploring how model prediction scores distributed over the dataset.}
According to the questionnaire, most model developers considered understanding the overall model behavior and performance \doc{an} 
essential step for analyzing NLP models.
As the final output of the model, prediction scores are a major signal \doc{of} 
its behavior~\cite{ren2016squares}.
To obtain \liushixia{an overview of the model performance}, \liushixia{$E_1$} to \liushixia{$E_5$} agreed that it was essential to show the distribution of prediction scores over the dataset.
For example, \liushixia{$E_1$} said, ``I would like to see whether the model makes mistakes on a particular set of similar samples or on diversified samples.''



The \textbf{\normalsize sample}-level requirements reflect users' need \doc{to identify} the samples of interest (F2) and \doc{analyze} them (F3) in a unified way. 

\emph{\normalsize R2. Identifying \doc{samples} that are essential for understanding and debugging the model.}
\liushixia{A common need expressed by the questionnaire participants is to find samples that are useful for model understanding and debugging} (F2).
Such samples of interest can be characterized from multiple aspects.
For example, checking the samples \doc{for} 
which the model makes a wrong prediction can help quickly debug the model (\liushixia{$E_1$--$E_5$}).
Investigating the samples that are close to the decision boundary can help increase model robustness~\zhen{~\cite{yang2020boundary}}.
Discovering the representative samples that are similar to
many samples may \shixia{shed} \doc{light on} 
why the model achieves \zhen{good or bad} performance.
Moreover, the experts required a way to identify the samples of \jw{interest} 
from the word \doc{perspective,} e.g., finding samples with a word that the model fails to correctly understand.\looseness=-1

\emph{\normalsize R3. Revealing how NLP models learn low- and high-level features of a sample across layers in a unified way.}
After identifying the samples of interest, the \liushixia{experts} needed to understand how the model processes the samples across layers and why it makes a certain prediction for the samples (F3, F6).
This allows them to figure out the underlying working mechanism of the model, which is important for model understanding and debugging.
Most existing tools help reveal important words (low-level features) in a sample. 
In addition to the low-level features, the \liushixia{experts} are also interested in the high-level features learned, e.g., whether the model can correctly understand sentence structures ($E_1$-$E_4$).
For example, \liushixia{$E_3$} said, 
``It is interesting to see whether a model judges the sentiment of a long compound sentence by considering the word relationships (e.g., phrases) or simply by counting positive and negative words.''
In addition, to facilitate model comparison, the \liushixia{experts} noted that it was necessary to provide consistent results for different models.

The \textbf{\normalsize word}-level requirements focus on identifying the words of interest (F4) and analyzing these words (F5) in a unified manner.

\emph{\normalsize R4. Identifying the words that are important for understanding and debugging the model.}
As an NLP dataset typically contains \xiting{tens of thousands of words},
it is very difficult for a user to check each word manually and decide which word is important for model understanding and debugging.
Accordingly, the \liushixia{experts} required a method to help them quickly identify the words of interest (F4), e.g., \xiting{ambiguous} words or words that contribute the most to model prediction (\liushixia{$E_1$-$E_5$}). 
 
\emph{\normalsize R5. Revealing how the model understands the meaning of a word by considering the context.}
After identifying the important words, the experts wanted to further investigate how the words affect the model prediction (F5), so that they can judge whether the model understands the meaning of the words in a correct way (\liushixia{$E_2$-$E_5$}).
Instead of considering each word independently, most NLP models consider a word by simultaneously modeling its context (the related words in the same sample).
To better understand how a word impacts a model, the \liushixia{experts} are interested in knowing more about the contextual information of the word.
For example, \liushixia{$E_4$} said, ``I would like to see whether the model can correctly distinguish different meanings of \emph{\normalsize like} based on other words in the sentences.''\looseness=-1

%% file: 4-visualization.tex
\section{DeepNLPVis}\label{sec:vis}
\subsection{Overview}

The large amount of NLP-model-related data, such as samples, words, and information-based measure data, \jw{makes} it difficult for users to identify the most informative information for model understanding.
To tackle this issue, we have developed a multi-level visualization and combined it with a unified information-based measure for understanding a deep NLP model from the perspectives of the corpus, samples, and words.

As shown in Fig.~\ref{fig:workflow}, the information-based measure, including intra-word information and inter-word information, is first extracted. 
Then based on the extracted measure, the three-level visualizations are seamlessly coordinated together and support an iterative \textbf{\normalsize analysis workflow} for a unified understanding of model training behaviors.
The \textbf{\normalsize corpus-level visualization} visually illustrates 1) the overall model performance \doc{of} all classes in a confusion matrix (class view); 2) the training samples of the two selected classes from the class view and the corresponding important keywords in a hexagonal heatmap (distribution view). This visualization enables quick identification of the classes, samples, and words of interest (\emph{\normalsize R1}, \emph{\normalsize R2}, \emph{\normalsize R4}).
Selected samples are displayed in the \textbf{\normalsize sample-level visualization}, and their associated words are displayed in the \textbf{\normalsize word-level visualization} for closer examination.
For example, users can identify the samples with wrong predictions and words with higher contribution values by using the sample list (\emph{\normalsize R2}) and word contribution view (\emph{\normalsize R4}).
Then for each sample, users can explore \jwR{the relationships between words and how they form and change across layers (e.g., formation of phrases) } in the information flow (\emph{\normalsize R3}).
They can also analyze the word of interest in the context of relevant words with the word context view (\emph{\normalsize R5}). 

\begin{figure}[!t]
\centering
\includegraphics[width=\linewidth]{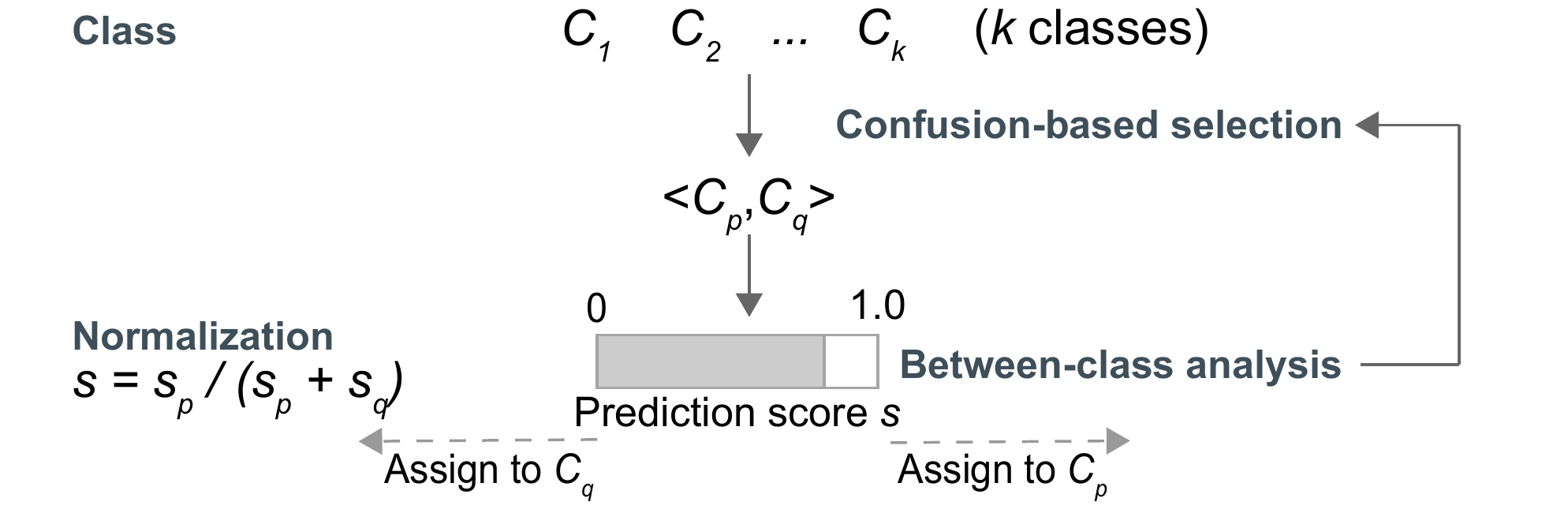}
\caption{
\xitingr{The analysis of multi-class classification is achieved by an iterative between-class analysis.}
}
\label{fig:multi-class}
\end{figure}


\subsection{Information-Based Interpretation}
\label{sec:information-based-sample-interpretation}

\xitingr{
We propose an information-based interpretation method for identifying the key information used by an NLP model for prediction. 
Without loss of generality, we introduce how to interpret a multi-class classifier.
When analyzing the performance of a classification model, the experts usually start with an overall class-level analysis, and then perform the between-class analysis. 
Inspired by this observation, we simplify the analysis of multi-class classification into an iterative between-class analysis, as illustrated in Fig.~\ref{fig:multi-class}.
In particular, the users first select two classes (e.g., $C_p$ and $C_q$), based on the confusion matrix for $k$ classes in the class view.
We then remove the impact of the unselected classes by computing} a normalized prediction score $s=s_p/(s_p+s_q)$, where \xitingr{$s_p$} and \xitingr{$s_q$} are the original prediction scores for \xitingr{$C_p$} and \xitingr{$C_q$}.
\xitingr{Normalized prediction score $s$ helps understand whether the model is confident or confused with respect to $C_p$ and $C_q$:
}
a large $s$ above $0.5$ indicates a confident prediction of \xitingr{$C_p$}, a small $s$ below $0.5$ indicate a confident prediction of \xitingr{$C_q$}, 
\xitingr{
and a value of $s$ around $0.5$ indicates that the model confuses the two classes in terms of the sample.
The goal of our interpretation method is to understand what information the model leverages for deciding the prediction $s$.
}

\begin{figure}[!t]
\centering
\includegraphics[width=\linewidth]{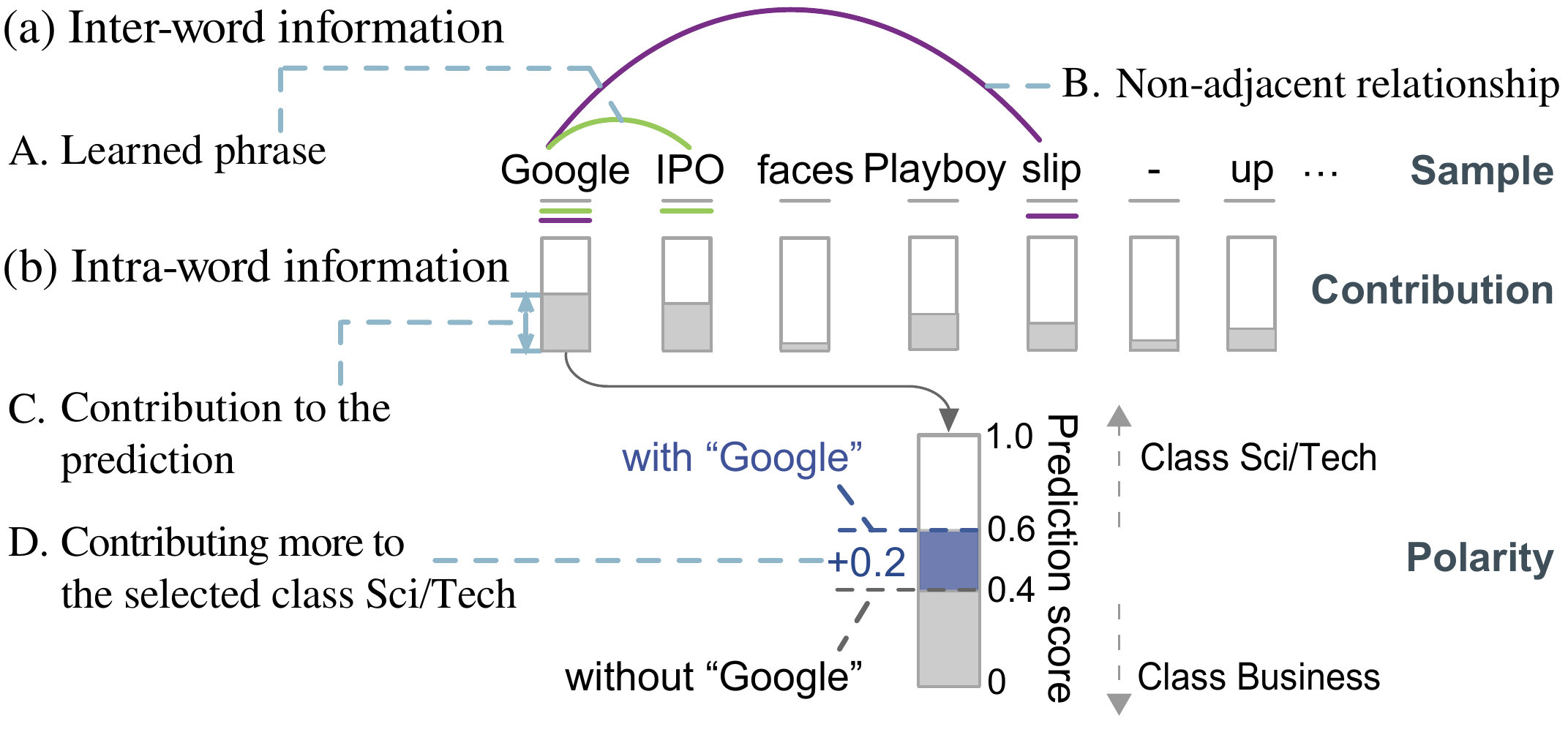}
\caption{
\jw{The visual explanation of intra- and inter-word information.}
}
\label{fig:explanation}
\end{figure}

\begin{figure*}[b]
\centering
\includegraphics[width=\textwidth]{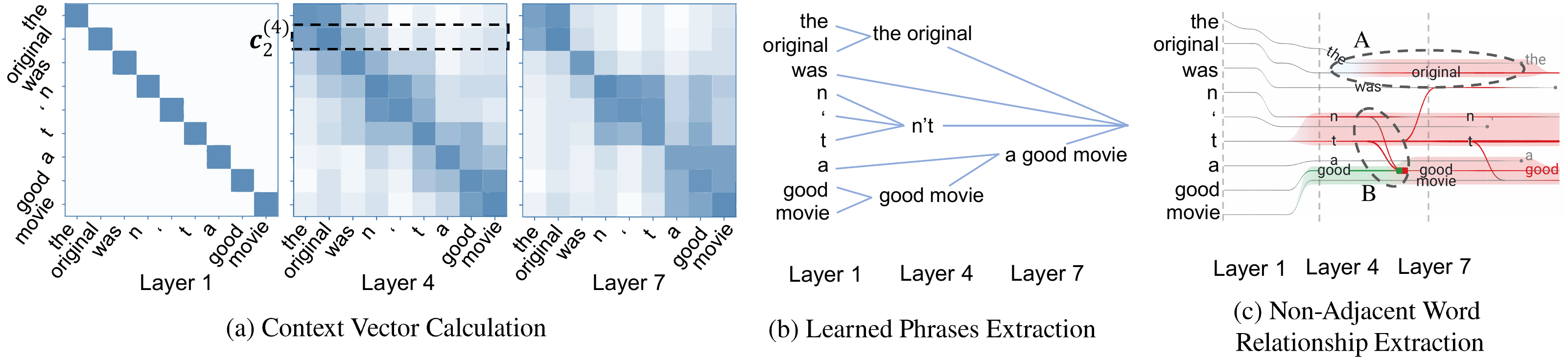}
\caption{Illustrating the learned phrases and non-adjacent word relationship of a sample. 
}
\label{fig:syntax_vis}
\end{figure*}

\xitingr{
The information is divided into two categories: \textbf{intra-word} and \textbf{inter-word}.
Intra-word information helps analyze how each single word contributes to the sample prediction, and inter-word information aims to capture the relationships between words (Fig.~\ref{fig:explanation}).
}
\xitingr{Recently,} 
Guan~\etal~\cite{guan2019towards} proposed a unified information-based measure to estimate \xitingr{the \textbf{contribution} of the word to the prediction (Fig.~\ref{fig:explanation}C), which partially addresses the problem of analyzing intra-word information}.
Given a sample ${\bf X}=(w_1,w_2...,w_n)$ and its prediction score $s$, the contribution of the $i$-th word $w_i$ at the $\ell$-th layer 
is measured \liushixia{by the amount of information that is passed from layer $l$ to the final prediction. 
It is computed as} mutual information: $\text{MI}(h^{(\ell)}(w_i);s)$,
where $h^{(\ell)}(w_i)$ is the latent representation of the $i$-th word of the sample at the $\ell$-th layer.
\xitingr{For models where there exist no 1-to-1 association between the latent representations and input words (e.g., CNN~\cite{kim2014convolutional}),  $h^{(\ell)}(w_i)$ is set to the concatenation of all hidden representations affected by $w_i$.}
The mutual information can be computed efficiently by using perturbation-based approximation~\cite{guan2019towards}.
The basic idea is to perturb $h^{(\ell)}(w_i)$ by adding a Gaussian noise \jwR{$\bm \epsilon_i$} and measure the magnitude of change in the prediction score $s$:
\begin{equation}
    \Delta s_i = \mathbb{E}_{{\bm \epsilon}_i\sim N({\bf 0},{\sigma^*_i} \textbf{I})} \frac{\phi(h^{(\ell)}(w_i)+{\bm \epsilon}_i)-s}{{\sigma}_s}.
    \label{eq:delta_s}
\end{equation}
Here, $\phi(\cdot)$ is the prediction function represented by the layers after $l$ and satisfies $s=\phi(h^{(\ell)}(w_i))$.
The perturbation ${\bm \epsilon}_i$ is a noise sampled from the Gaussian distribution $N({\bf 0},{\sigma^*_i} \textbf{I})$, where $\sigma^*_i$ is the optimal standard deviation computed by using the maximum likelihood estimation loss~\cite{guan2019towards}, $\textbf{I}$ is an identity matrix, and ${\sigma}_s$ is the standard deviation of $s$.
A larger magnitude change $\lvert \Delta s_i\rvert$ indicates a larger contribution of the word (Fig.~\ref{fig:explanation}C).

Although this measure can compute the absolute value of word contribution, 
it fails to provide information for understanding the polarity of the contribution. 
Moreover, it fails to capture the relationships between words (inter-word information).
Here, the \textbf{\normalsize polarity} indicates towards which class the word or word combination (e.g., phrase) contributes to the prediction, given two classes selected from the class view \xitingr{(Fig.~\ref{fig:explanation}D)}. 
\xitingr{
The polarity is important for identifying the root cause for the confusion between \grammarly{the} two classes.
For example, the sample in Fig.~\ref{fig:explanation} is misclassified to ``sci/tech" because the model considers ``google" as a word related to ``sci/tech'' rather than ``business'', even though ``google" is mentioned together with ``ipo."
}
\xitingr{In addition to polarity, another important type of information is inter-word information, which} is useful for detecting high-level features learned by the model, such as phrases \xitingr{(Fig.~\ref{fig:explanation}A)} and non-adjacent word relationships \xitingr{(Fig.~\ref{fig:explanation}B)}.
\xitingr{Next, we introduce} how we extend the information-based measure to learn the polarity of the contribution and inter-word information.

The polarity of the word is measured by the sign of change in the prediction score with the existence of that word (Fig.~\ref{fig:explanation}D).
Specifically,
\xitingr{$\Delta s_i>0$ means that removing the $i$-th word increases the prediction score $s$, which indicates the existence of $w_i$ contributes to assigning a sample to class $C_q$.} 
Thus, $w_i$ is a \xitingr{$C_q$-relevant word}.  
$\Delta s_i<0$ means that \jwR{$w_i$ is a \xitingr{$C_p$-relevant word}.} 
To increase the robustness of the method, we further \shixia{use} a margin $\xi>0$ to extract the most \xiting{relevant} words.
$\Delta s_i>\xi$, $\Delta s_i<-\xi$, or $-\xi\leq\Delta s_i\leq\xi$ \xitingr{mean that word $w_i$ is a $C_q$-relevant word, $C_p$-relevant word, and class-irrelevant word.
A good value of $\xi$ should well differentiate class-relevant and -irrelevant words.
For example, words like ``ipo" and ``spending" should be considered relevant with ``business," words like ``search" and ``gates" should be considered relevant with ``sci/tech," and words like ``time" and ``can" should be class-irrelevant.
We experiment with seven datasets and find that the} best \xitingr{value for $\xi$ usually increases with increasing text length (see supplement for detailed results).
We suspect that this is because for longer text, the information is scattered across more words, resulting in a larger variance of mutual information and thus the requirement for a larger margin $\xi$. Since the experts usually use datasets with short texts, we set the value of $\xi$ in the system to 0.02, which typically works well for short text whose average number of characters is smaller than 300. 
We also allow users to interactively change the value of $\xi$ for a given dataset.
}

The inter-word information reveals how an NLP model models the phrases \jwR{in} 
a sample and learns the relationships between non-adjacent words (Fig.~\ref{fig:explanation}(a)).
Although different types of neural networks model word relationships \jw{in different ways,} 
(e.g., Transformer uses self-attention and CNN leverages convolutional kernels), they all embed the learned relationships into the contextual word embedding~\cite{tenney2019you}.
For word $w_i$, each model identifies its most relevant words (context) and \grammarly{encodes} \liushixia{them} into the latent representation $h^{(\ell)}(w_i)$.  
Based on this, we probe into the learned phrases and non-adjacent relationships by analyzing the word information contained in $h^{(\ell)}(w_i)$.
Our method for extracting the inter-word information consists of three steps:

\noindent\underline{\emph{\normalsize Step 1. Computing context vector.}}
\xitingr{
The context vectors of words help identify word clusters (phrases) based on the information each word absorbs.
For example, at layer 1, each word \doc{only contains} information about itself (Fig.~\ref{fig:syntax_vis}(a)).
}
Later, the words (e.g., ``good'' and ``movie'') absorb information from each other, and their context vectors become more similar (layers 4 and 7 in Fig.~\ref{fig:syntax_vis}(a)).
\liushixia{The context vector ${\bf c}_i^{(\ell)}$ of word $w_i$ is created by decomposing the word information contained in $h^{(\ell)}(w_i)$: $c^{(\ell)}_{ij}= MI(w_j, h^{(\ell)}(w_i))$.}
Here, $c^{(\ell)}_{ij}$ reveals how much information of the $j$-th input word is used in the contextual word embedding of the $i$-th word.

\noindent\underline{\emph{\normalsize Step 2. Extracting the learned phrases.}}
The learned phrases are extracted by clustering the context vectors (Fig.~\ref{fig:syntax_vis}(b)).
We employ the agglomerative clustering~\cite{rokach2005clustering} to cluster adjacent context vectors at each layer.
To improve stability, the clustering result of layer $\ell$ is utilized to initialize the clusters at layer $\ell+1$.
As shown in Fig.~\ref{fig:syntax_vis}(b), the clustering allows us to find phrases extracted by the model, e.g., ``the original'' at layer 4 and ``a good movie'' at layer 7.

\noindent\underline{\emph{\normalsize Step 3. Extracting non-adjacent word relationships.}}
\shixia{In addition to showing how adjacent words form phrases, 
we also consider} the interactions between non-adjacent words.
\jw{Taking} the sample in Fig.~\ref{fig:syntax_vis}(c) as an example, ``n't'' interacts with ``good'' even though they are non-adjacent.
This interaction can be quantified by using mutual information:
$e^{(\ell)}_{ij}$=$MI(h^{(\ell)}(w_i);h^{(\ell+1)}(w_j))$, where $i$ and $j$ are the indices of two non-adjacent words at layer $\ell$.
\xiting{A large $e^{(\ell)}_{ij}$ reveals that \doc{a great deal of} 
information has been passed from the $i$-th word to the $j$-th word at \jwrevision{layer} $\ell+1$.}
\looseness=-1

\subsection{Three-Level Visualization}
The three-level visualization enables users to smoothly navigate from the overall performance at the corpus-level to the detailed information at the sample- and word-level.

\subsubsection{Corpus-Level Visualization}
\label{sec:corpus-level-visualization}
The corpus-level visualization contains two parts: a \textbf{\normalsize class view} to reveal the overall model performance on all classes and a \textbf{\normalsize distribution view} to explore the prediction distribution over the selected classes from the class view and identify important samples and words for further analysis.

\shixia{A confusion matrix is employed in the class view (Fig.~\ref{fig:teaser}(a)), 
where each column} represents the samples in a predicted class, each row represents the samples in an actual class, and the percentage of samples displayed in each cell depicts the confusion between two classes.\looseness=-1 

For the distribution view, the Squares visualization~\cite{ren2016squares} is a straightforward solution to visually convey the desired information.
Although Squares can well show the model performance in the context of samples, it fails to disclose the similarity relationships between samples. 
Understanding such relationships are critical for identifying important samples, such as representative samples and outliers in each class, for further investigation.

To tackle this issue, we have developed a hexagonal heatmap (Fig.~\ref{fig:teaser}(b)), which integrates the prediction score (y-axis) with the \doc{one-dimensional} t-SNE projection~\cite{van2014accelerating}. 
\liushixia{Since the users require to examine the similarity relationships between samples, we employ the t-SNE projection.
We choose this technique because of its effectiveness in preserving the neighborhoods and clusters of samples~\cite{maaten2008visualizing}.}
The one-dimensional t-SNE projection projects the sample embedding in the last hidden layer of the model onto \doc{the} x-axis. 
The sample color encodes the class of the sample.
\liushixia{For example, in Fig.~\ref{fig:teaser}}, \zhennew{orange and blue encode the ``business" and ``sci/tech" news}, respectively.
In the hexagonal heatmap, a blue hexagon \includegraphics[height=\fontcharht\font`\B]{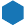} represents a set of true-positive samples, an orange hexagon \includegraphics[height=\fontcharht\font`\B]{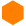} represents a set of true-negative samples, a blue cross \includegraphics[height=\fontcharht\font`\B]{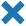} represents a set of false-negative samples, and an orange cross \includegraphics[height=\fontcharht\font`\B]{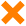} represents a set of false-positive samples. 
The darker the sample color of a hexagon/cross is, the more samples it represents. 
To provide a comprehensive overview of how samples distribute over prediction scores (y-axis), we bin them into 16 consecutive stripes with an interval of \zhen{0.0625}.
Each stripe contains a pie chart on the right to illustrate \xiting{the class distribution over the samples}.\looseness=-1

A group of representative words is placed on the other side in a layout \doc{close} to \xitingr{the samples}.
\xiting{
Different colors represent words that are relevant to different classes, and gray represents class-irrelevant words. 
} 
The size of each word encodes its importance to the model prediction. 
In NLP, the term frequency–inverse document frequency (TF-IDF) weighting scheme is widely used to measure how important a word is to a \emph{\normalsize document}~\cite{yahav2018comments}. 
This weighting scheme assigns higher weights to the words whose occurrence is frequent in a small number of documents, but rare in the other documents of the corpus. 
Inspired by TF-IDF, we compute $\normalsize \text{Importance}(w)$, which measures how important a word is to a \emph{\normalsize model} in a \emph{\normalsize corpus} by \jwR{1) }its term frequency $\text{tf}(w)$ in the corpus; and 2) its average contribution over the associated samples, $\normalsize \text{Contribution}(w)$:
\begin{equation}
   \normalsize \text{Importance}(w) = \log(\text{tf}(w)+1.0) * \text{Contribution}(w),
    \label{eq:importance}
\end{equation}
where the first term is the term frequency of word $w$ in the document collection. \jwR{As the $\text{tf}$ values span a large range, a logarithmic 
operation is applied to normalize the frequency values.} 
All the tf values are also increased by 1 in the $\log$ normalization to avoid zero output. 
The second term measures the contribution of $w$ to the model prediction.
Accordingly, a large importance value is attained by a high term frequency and a high contribution score. 
Such a weighting method tends to filter out common words with little contribution to the final prediction or rare words with low frequency.\looseness=-1


To explore the overall model behavior at the corpus level, rich interactions \shixia{are} designed. 
For example, when a user finds a stripe of interest, 
s/he can click \shixia{to enlarge} the stripe and \zhen{examine at a finer scale with more keywords}. 






\subsubsection{Sample-Level Visualization}
The sample-level visualization consists of two coordinated components: a sample list and an information flow (Fig.~\ref{fig:teaser}).

The sample list (Fig.~\ref{fig:teaser}(d)) allows users to examine multiple samples in terms of their text content, class labels, and prediction scores.
Sorting \shixia{by} these attributes makes it possible to identify the samples of interest from multiple \doc{perspectives} (\emph{\normalsize R2}).
The list is coordinated with other visualizations. 
For example, the samples in the list will be updated according to the selected sample hexagons or words in the corpus- or word-level visualizations. 
Users can choose a sample from the list and perform a deeper analysis of the word relationships and their changes by using the information flow.\looseness=-1

The information flow (Fig.~\ref{fig:teaser}(e)) facilitates users to analyze how NLP models process a sample through layers in a unified way 
(\emph{\normalsize R3}). 
\liushixia{The design is inspired by storyline visualization~\cite{tanahashi2012design}}.
\liushixia{We introduce the visual encoding and layout algorithm below.}

\noindent \textbf{\normalsize Visual encoding.}
As shown in Fig.~\ref{fig:syntax_vis}(c), in the flow visualization, each word is represented by a line.
The \textbf{\normalsize intra-word} information is encoded by the line width and colors.
A wider line indicates that the word contributes more to the final prediction.
\xitingr{The color of the line indicates which class the word contributes to (polarity).}
Additionally, we highlight two types of important changes \shixia{along} a line. 
\shixia{First}, if a word contributes \jwR{little} 
to the final prediction after layer $\ell$, we will end the line with \includegraphics[height=\fontcharht\font`\B]{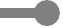}. 
\shixia{Second,} we highlight the \shixia{class} change on a line by using glyphs \includegraphics[height=\fontcharht\font`\B]{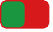} and \includegraphics[height=\fontcharht\font`\B]{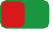}. 
The \textbf{\normalsize inter-word} information includes the learned phrases and non-adjacent word \doc{relationships}.
For the learned phrases, we use the distances between lines to represent the distances between word context vectors.
In this way, words that belong to a phrase are naturally placed \doc{close} 
to each other.
The clusters are further highlighted by using a background area (Fig.~\ref{fig:syntax_vis}A), \doc{the color of which} 
is determined by the \doc{primary} 
color of the lines.
The \doc{relationships} 
between non-adjacent words are encoded by the curves that connect different lines (Fig.~\ref{fig:syntax_vis}B).
The width of a curve from word $i$ at layer $\ell$ to word $j$ at layer $\ell+1$ is determined by mutual information $e^{(\ell)}_{ij}$ (Sec.~\ref{sec:information-based-sample-interpretation}).
A thicker curve indicates a larger contribution from word $i$ \jw{to} word $j$.
To avoid visual clutter, we only display the most important curves.
A curve is shown if 1) its weight $e^{(\ell)}_{ij}$ is among the top 5\%,
or 2) it is useful for illustrating the color/width change of a line.
For example, the curve displayed in Fig.~\ref{fig:syntax_vis}B is helpful for explaining why \doc{the} word ``good'' becomes negative at layer 6, which is caused by “good” absorbing the information of \xiting{``n't.''}

\noindent \textbf{\normalsize Layout algorithm.}
The layout of the storyline needs to preserve both stability and readability~\cite{tanahashi2012design,liu2013storyflow}.
To achieve this goal, we formulate the layout as a constrained optimization problem.

Denote the y-coordinate of word $i$ at layer $\ell$ as $y^{(\ell)}_i$.
\textbf{\normalsize Stability} prevents the y-coordinate of a line from changing dramatically when its context vector does not change much.
This ensures that the \jw{wiggles of the line, which easily draw users' attention,} are meaningful and worth 
\lizhen{investigating.}
We consider two types of stability losses: a continuous loss $(y^{(\ell)}_i-y^{(\ell-1)}_i)^2$ and a discrete loss $\mathbb{I}(y^{(\ell)}_i\neq y^{(\ell-1)}_i)$.
$\mathbb{I}(\cdot)$ is an indicator function with $\mathbb{I}(\text{true})=1$ and $\mathbb{I}(\text{false})=0$.
While the continuous loss \jwR{penalizes} 
large changes, the discrete loss \jwR{limits} 
the number of line wiggles.
\textbf{\normalsize Readability} 
measures how clear and easy \doc{it is} to understand the \liushixia{relationships between words in a sample.}
In addition to the phrase relationships between adjacent words, \jwR{the order of words} is also important for many NLP tasks.
For example, \emph{\normalsize isn't he lovely} and \emph{\normalsize he isn't lovely} have different \jw{sentiments.} 
As a result, readability requires that 1) the distances between lines accurately reveal the distances between word context vectors; and 2) \jwR{the order of words} 
in a sample is preserved.
Accordingly, readability is maintained by \jwR{minimizing the loss} 
$(||y^{(\ell)}_i-y^{(\ell)}_{i-1}||-D^{(\ell)}_{i})^2$ and \jwR{satisfying the constraint} 
$y^{(\ell)}_i\geq y^{(\ell)}_{i-1}$, for $\forall i,l$, where $D^{(\ell)}_{i}=||{\bf c}^{(\ell)}_i-{\bf c}^{(\ell)}_{i-1}||$ is the distance between word context vectors.

\shixia{Based on the analysis of stability and readability, we formulate the} storyline layout as a constrained optimization problem:
\begin{equation}
    \begin{aligned}
     \min_{\{y^{\ell}_i|\forall i,\ell\}} & \ \sum_{i=1}^M \sum_{\ell=1}^L C(i,\ell), \ \ \ \ s.t., y^{(\ell)}_i\geq y^{(\ell)}_{i-1},\ \ \forall i,\ \ell \\
     C(i,l) &=\alpha[(y^{(\ell)}_i-y^{(\ell-1)}_i)^2+\beta \mathbb{I}(y^{(\ell)}_i\neq y^{(\ell-1)}_i)]\\
    & +(1-\alpha)(||y^{(\ell)}_i-y^{(\ell)}_{i-1}||-D^{(\ell)}_{i})^2
    \label{eq:storyline_layout_problem}
    \end{aligned}
\end{equation}
The first two terms of $C(i,l)$ maintain stability, and the third term \jwR{maintains} 
readability.
$M$ is the number of words in a sample, and $L$ is the number of layers in the NLP model. $\alpha \in [0, 1],\ \beta>0$ are hyperparameters that balance different terms in the loss.
In our implementation,  \zhennew{$\alpha = 0.4$}, and $\beta = 5$.

The constrained problem defined in Eq.~(\ref{eq:storyline_layout_problem}) can be solved by dynamic programming in pseudopolynomial time.

\subsubsection{Word-Level Visualization}
The word-level visualization consists of a \textbf{\normalsize word contribution view} and a \textbf{\normalsize word context view}.

\noindent\textbf{Word contribution view}.
The word contribution view helps identify the words and layers of interest  
based on the words' contribution to different layers in the model (\emph{\normalsize R4}).

\textit{\normalsize Visual encoding}.
As shown in Fig.~\ref{fig:teaser}(c), the x-axis denotes layers, and the y-axis corresponds to the word contribution percentiles.
\liushixia{Words are divided into 10 equal-size groups based on their contribution at each layer.}
The groups with larger contributions are placed higher on the $y$-axis. 
In this way, the visualization reveals whether the model leverages the correct words for prediction.
For example, we can debug a \zhennew{news classification} model by checking whether \zhennew{class-relevant words like ``antitrust" and ``google"} are placed on the top \liushixia{at the last few layers}, and other words like \zhennew{``this," and ``has"} are placed at the bottom.

Following the corpus-level visualization, 
the size of a word encodes its importance.
The color of a word is determined by its dominant polarity in its associated samples. 
A pie-chart-based glyph \includegraphics[height=1.2\fontcharht\font`\B]{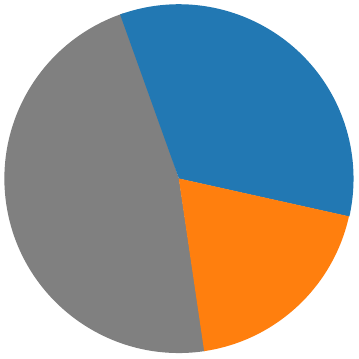} is used to show the distribution of word polarity in all the associated samples.
To reduce visual clutter, we only show pie charts for the top five most important words at each layer.

\textit{\normalsize Layout}.
The words are placed based on the sweepline algorithm~\cite{shi2010understanding},
which places important words close to the centroid of a given contour.
We slightly modify the algorithm by placing each word $w$ close to its desired position. 
This is achieved by replacing the centroid with  $(\tilde{x}^{(\ell)}_w,\tilde{y}^{(\ell)}_w)$.
Here, $\tilde{y}^{(\ell)}_w$ is the \xiting{contribution percentile} of $w$.
To ensure stability, we try to maintain the relative position of $\tilde{x}^{(\ell)}_w$. 
If the word usually appears at the left (or right) side of previous layers, we prefer to place it at the left (or right) side of the current layer.  

\begin{figure*}[t]
\centering
\includegraphics[width=1\textwidth]{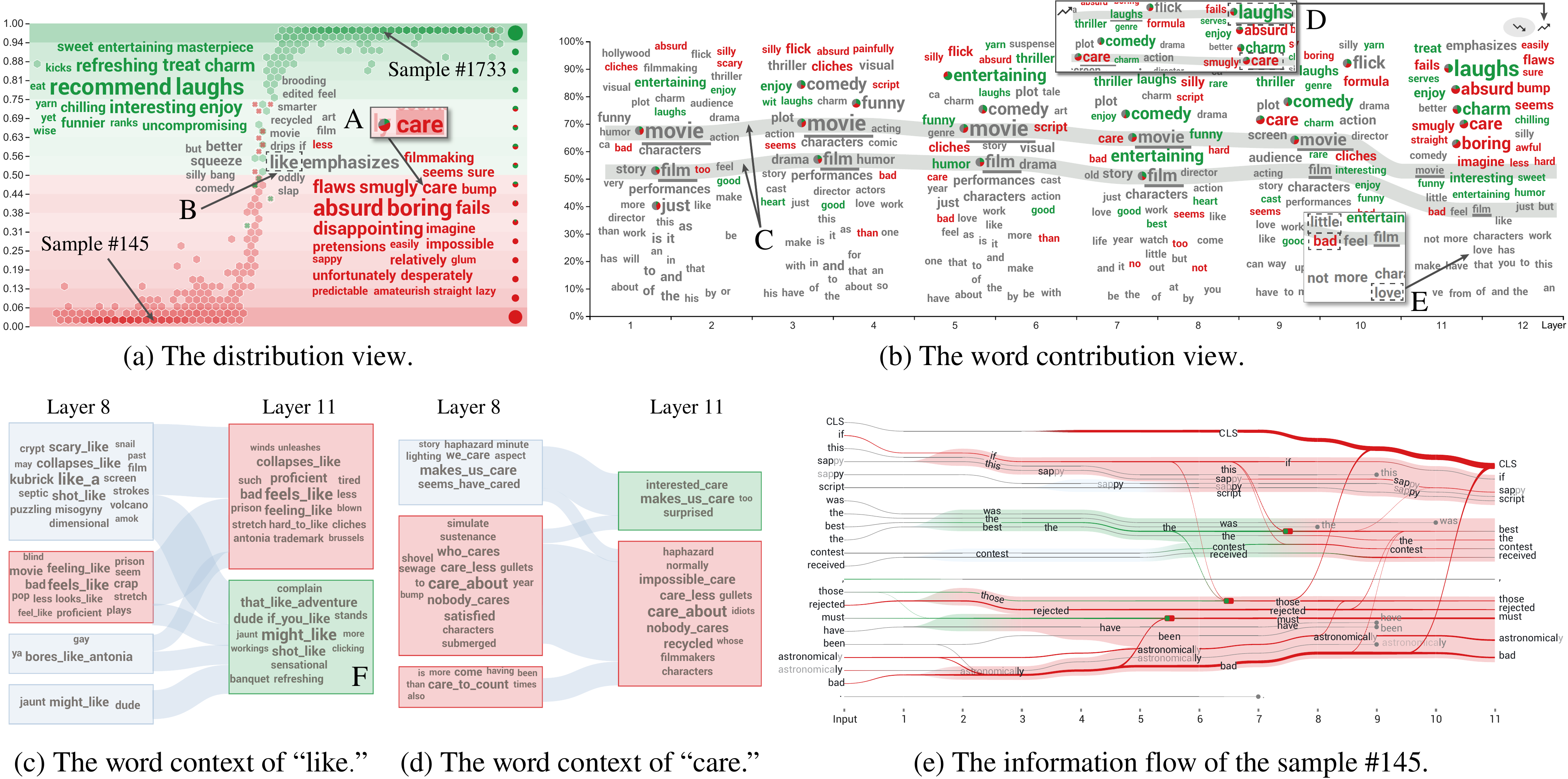}
\caption{
The analysis of the BERT model on sentiment classification.
}
\label{fig:sentiment-bert}
\end{figure*}

\textit{\normalsize Interaction}.
This view is coordinated with other views to understand the model from the word perspective. For example, 
the words displayed will be updated upon the selection of samples or words in other views.
We can hover \doc{over} a word to inspect its polarity distribution over samples \shixia{with a pie chart} and highlight its appearances across layers.
To help identify interesting words, we enable two types of automatic pattern searching functions based on the the experts' suggestions.
Trending button \includegraphics[height=\fontcharht\font`\B]{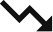} (or \includegraphics[height=\fontcharht\font`\B]{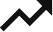}) in the top right corner is used to show the words whose contributions keep decreasing (or increasing) through layers 
(Fig.~\ref{fig:sentiment-bert}C \shixia{or} D).

\noindent\textbf{Word context view}.
The word context view facilitates \jwR{the understanding of a word} by revealing how the model processes it based on its context (\emph{\normalsize R5}).

The context of word $w$ is depicted by a list of words that are considered \doc{to be} the most relevant to it \doc{according to} 
the NLP model.
The word context view illustrates the context of $w$ across different \emph{\normalsize samples} and \emph{\normalsize layers}.
Fig.~\ref{fig:sentiment-bert}(c) shows an example context visualization of \doc{the} word ``like."
Each rectangle represents a cluster of samples
with similar context words for ``like."
Its size encodes the number of samples in the cluster, and its color encodes the majority polarity of ``like" in these samples.
For example, the meaning of ``like" in cluster F at layer 11 is ``favor," and its sentiment is mostly positive (e.g., ``if you like" and ``might like").
Phrases, such as 
``if you like'' are extracted by identifying words that are both relevant and adjacent to ``like'' in the samples. 
The font size of the phrase is used to encode the importance score. 
The sample clusters are computed by performing agglomerative clustering~\cite{rokach2005clustering} on the word context vectors (Fig.~\ref{fig:syntax_vis}(a)).  
We initialize the clusters at each layer by using the clusters extracted in the previous layer to \doc{maintain} stability.
The cluster positions are determined by using the \shixia{directed acyclic graph} layout algorithm employed in TextFlow~\cite{cui2011textflow}.
The width of the edge encodes the proportion of samples that come from the previous cluster.
To reduce visual clutter, we only show the results of \shixia{representative layers, 
whose similarity with the} previously selected representative layer is \zhen{smaller} than a threshold.

The word context view is coordinated with other views
\jwR{during the analysis.} 
\jwR{It is triggered when a user selects a word in the distribution view or the word contribution view. By selecting a rectangle in the word context view, the corresponding samples are highlighted in the distribution view and sample list as well.} 


%% file: 5-application.tex
\section{Case Studies}
\label{sec:application}

\doc{We conducted case studies involving three tasks with experts \zhennew{$E_3$} and \zhennew{$E_4$}.} 
\liushixia{$E_3$ is interested in understanding and diagnosing BERT models for binary/multi-class classification tasks,
and \liushixia{$E_4$} would like to compare different models using DeepNLPVis.}

In conducting the case studies, DeepNLPVis requires the inter- and intra-word information for each sample in the training set. The calculation for the whole set is time-consuming and is thus carried out offline. 
All the other \liushixia{required data} can be obtained with real-time processing (within one sec), and are thus calculated online.

\subsection{Binary Sentiment Classification}
In this case study, \liushixia{$E_3$} carried out the sentiment classification task on the Stanford Sentiment Treebank (SST-2)~\cite{socher2013recursive}, 
which consists of sentences from movie reviews and human annotations of their sentiments. \jwrevision{The sentiments are of two classes: positive and negative.} 
The GLUE SST-2 splits~\cite{wang2019glue} were used for the training (67,349 samples), validation (872 samples), and test (1,821 samples) sets.
The BERT model, as the most widely used NLP model, was applied as the baseline and achieved 93.23\% accuracy.
Starting with the BERT model, \liushixia{$E_3$} emphasized on gaining \shixia{a} comprehensive understanding of the model's working mechanism, which would in turn facilitate the subsequent model diagnosis.

\subsubsection{Understanding}

\noindent\textbf{\normalsize Understand the overall performance} (R1, R4). 
\liushixia{$E_3$} began the analysis by examining how the prediction scores \doc{were} distributed over the data in the \lizhen{distribution} view (R1). 
He immediately noticed the long-tailed distributions (Fig.~\ref{fig:sentiment-bert}(a)), 
showing that most samples \doc{were} 
predicted with high confidence (vertically away from the center), 
while those with low confidence \doc{were} horizontally closer to the center.
\liushixia{$E_3$} also noticed the keywords extracted from the positive and negative samples \doc{had} relevant sentiments. 
The overall distribution \doc{in} the \lizhen{distribution} view \doc{gave} \liushixia{$E_3$} confidence \doc{in} the model's performance.
He then turned his attention to the word contribution view (Fig.~\ref{fig:sentiment-bert}(b)) to analyze how the model understands the words at different layers (R4). 
\jwrevision{He noticed that with the layers going deeper, the model recognized more words with sentiment tendencies, i.e., more colored words. Moreover, at layers 11 and 12, words with strong positive or negative meanings (e.g., ``absurd,'' ``laughs,'' ``charm'') contributed highly. These sentimental words replaced those class-irrelevant ones (e.g., ``flick,'' ``screen,'' ``filmmaking'') in previous layers.}  

\liu{Analyzing how the contributions of words changed through the layers (R4) is helpful to understand the model's working mechanism.
} 
Thus, \liushixia{$E_3$} clicked the trending hint buttons 
to display the top two words with the largest \jw{contribution} 
changes (decreasing or increasing). He noticed that the \jw{contributions} 
of ``movie'' and ``film'' decreased slowly until layer 10 and then decreased rapidly (Fig.~\ref{fig:sentiment-bert}C). While the \jw{contributions} 
of  ``laughs'' and ``care'' increased rapidly in the last few layers (Fig.~\ref{fig:sentiment-bert}D).
This observation \doc{verifies} 
his hypothesis that the BERT model, as a fine-tuned model, has 
the early layers more dedicated to learning transferable representations of language that are invariant to the prediction task, while the deeper layers \jwR{paying} 
more attention to words that are relevant to the prediction task~\cite{brunner2019identifiability}.

The word contribution view gave \liushixia{$E_3$} an \shixia{overview} of how the model gradually \doc{adapted} 
to the tasks through the layers.
To get a deeper understanding of the model, \zhennew{$E_3$} then \shixia{analyzed} how it \shixia{understood} 
the meaning of individual words and \shixia{processed} individual samples.

\noindent\textbf{\normalsize Understand words in context} (R5).
Some keywords in the distribution view aroused \zhennew{$E_3$}'s attention. He noticed that the word ``care'' (Fig.~\ref{fig:sentiment-bert}A) \jw{contributed} 
more to negative predictions that \doc{went} 
against his intuition. He thus examined its word context view (Fig.~\ref{fig:sentiment-bert}(d)). 
He observed that when the samples \jw{went} through the layers, the model \jw{tended} to embed ``care'' in a negative context , 
such as ``care less,'' ``nobody cares,'' ``who cares.''
Only a small \doc{number} 
were embedded in \doc{a} positive context. 
This \liu{is due to} the more frequent presence of ``care'' in the negative samples (\zhennew{191} negative and \zhennew{100} positive). 
Considering the SST2 data came from movie reviews, 
\liushixia{$E_3$} \shixia{thought it was} reasonable because ``care'' is indeed found more often in negative reviews.

\zhennew{$E_3$} further noticed the word ``like'' that appeared to be \jwrevision{class-irrelevant} 
rather than positive 
(Fig.~\ref{fig:sentiment-bert}B). 
Examining its word context view (Fig.~\ref{fig:sentiment-bert}(c)), \doc{there were} 
four main clusters formed at layer 8. 
\jwrevision{Three of them were class-irrelevant, i.e., non-apparent sentiment tendency at this layer, while one of them was negative. The negative cluster contained ``feels like,'' ``looks like,'' where ``like'' was a proposition with no strong sentiment tendency. However, the negative context in this cluster, such as ``crap,'' ``blind,'' and ``criticizing,'' associated this cluster with a negative sentiment.} 
\jwrevision{Then through the further exchange of context, the \jwrevision{class-irrelevant} 
clusters split. Parts of them fused into the negative cluster, while other parts formed a positive cluster at layer 11. In the positive cluster, ``like'' appeared more in phrases such as ``if you like,'' and ``might like'' with the meaning of ``favor.''} 
From the word context view, \liushixia{$E_3$} \shixia{was} more confident \doc{in} 
the BERT model's ability to disambiguate words with multiple meanings.\looseness=-1

\noindent\textbf{\normalsize Understand the prediction of samples} (R3).
To understand how the model processes a sample across layers,
\liushixia{$E_3$} selected the samples of interest by coordinating the \lizhen{distribution} view and sample list.
He first selected a set of samples with the highest confidence in the \lizhen{distribution} view and then examined their content in the sample list \jwR{to find samples with interesting structures}.
By repeating this step, he finally selected samples \#145 and \#1733 (Fig.~\ref{fig:sentiment-bert}(a)).
 They are predicted as negative and positive with turning structures.

Fig.~\ref{fig:sentiment-bert}(e) shows the information flow for \liushixia{predicting} the negative sample \#145 ``if this sappy script was the best the contest received, those reject must be astronomically bad.''
This sample is of a turning structure with the word ``if.'' 
At layer 1, the phrase ``contest received'' \doc{was} initially formed with \jwrevision{no-apparent sentiment tendency}. 
Then at layer 6, ``was the best the contest received'' \doc{was} formed together and regarded \doc{as} positive due to the positive sentiment of ``best.'' However, it \doc{turned} into negative at layer 7 where the information from ``if'' \doc{was} transferred to it, which indicates that the model correctly \doc{interpreted} the turning structure.
\jwrevision{Before layer 6, the information was transferred more locally such as from `astronomically bad'' to ``must,'' changing its sentiment into negative. At layers 6 and 7, the information from ``if'' \doc{was} transferred to the second half of the sentence. }
So far the model \doc{recognized} the overall structure, and then at layers 7 and 10, it \doc{transferred} the information from \jwrevision{the second half of the sentence} 
to \zhengyan{[CLS]}, which \doc{ultimately determined} the final negative prediction of this sample. 

From this negative sample, \liushixia{$E_3$} also gained insights \doc{into} 
how the model understands \jwR{a sentence}.
\jwrevision{At the beginning layers, words and phrases were formed, such as ``contest received,'' ``was the best,'' and ``astronomically bad''. 
Along with the layers going deeper, more sentiment information transfers were observed between words and phrases. It shows that the model was mostly devoted to understanding the sentence structure at early layers, while shifted its attention to transferring information relevant to the prediction task.}\looseness=-1


\zhennew{$E_3$} had similar observations for the positive sample \#1733 ``though the film is static, its writer-director’s heart is in the right place, his plea for democracy and civic action laudable.'' Again, this sample has a turning structure with the word ``though,'' which was recognized at layer 8. 
The recognition changed the sentiment of ``the film is static'' to positive, which, together with the positive sentiment of ``is in the right place'' and ``laudable'' 
\doc{resulted} 
in the final positive prediction.

\begin{figure}[t]
\centering
\includegraphics[width=1\linewidth]{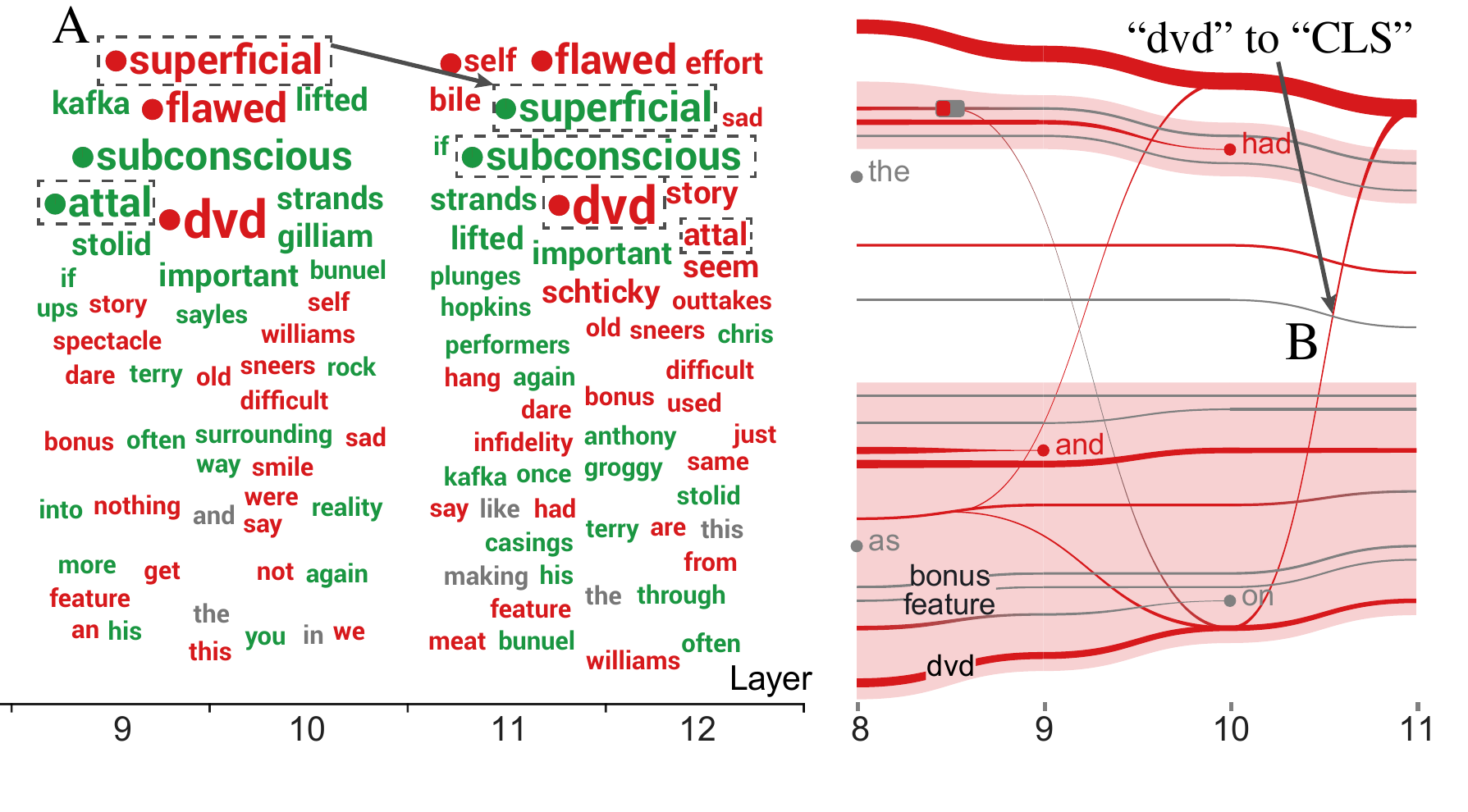}
{\rmfamily
\put(-207,-3){\textls[-5]{\spaceskip=0.16em\relax(a) Word-level.}}
\put(-82,-3){\textls[-5]{\spaceskip=0.16em\relax(b) Sample-level.}}
}
\caption{
\jw{Diagnosis of misclassified samples.
}
}
\label{fig:bert-diagnosis}
\end{figure}

\subsubsection{Diagnosis}
With a deeper understanding of how the model understands words and samples, \liushixia{$E_3$} then attempted to identify the deficiencies in the current model and \jwR{improve its performance} 
(R2).



He selected all the misclassified samples in the \lizhen{distribution} view (the red and green crosses in Fig.~\ref{fig:sentiment-bert}(a)) and turned to the word contribution view \zhennew{(Fig.~\ref{fig:bert-diagnosis}(a))} to examine the important words for predicting them.
He first noticed \lizhen{the} word ``superficial'' (Fig.~\ref{fig:bert-diagnosis}A), which remains strongly negative but turns to positive in the last layer. Wondering why, he clicked on it to examine the associated samples in the sample list and found that there \doc{was a conflicting label} 
for ``a superficial way'' (positive) and ``the superficial way'' (negative). 
As a fine-tuned model, the last few layers of BERT are more influenced by the training samples. 
\liushixia{$E_3$} considered the \doc{conflicting label confused the model} about the sentiment of ``superficial'' in the last layer and \shixia{resulted} in some of the misclassifications. 
He then corrected the \doc{conflicting} labels for the two samples.

Another word \doc{that drew} his attention \doc{was} ``dvd,'' which \doc{was} 
an important word through the layers. 
However, it \jwR{was strongly negative,} which \doc{went} against his intuition. 
Checking the associated samples, he found that there \shixia{were} far more negative than positive samples that \doc{included} ``dvd'' (39 negative and 14 positive).
\jwrevision{$E_3$ thus considered this was a case of the shortcut issue~\cite{mccoy2019shortcut} caused by the limited diversity of training data. Shortcut refers to the phenomenon that a model learns spurious correlations between words and labels, e.g., classifying a word to be positive/negative only based on its occurrence in according samples rather than understanding its inter/intra-word relationships. To see the influence to prediction, $E_3$ then turned to an individual sample to check the information flow.} 
He selected sample \#76, which is a positive sample but wrongly predicted as negative. 
The information flow of this sample (Fig.~\ref{fig:bert-diagnosis}(b)) showed that ``dvd'' was regarded as a negative word from the beginning and throughout the layers. \jwrevision{Its negativity was passed to [CLS] at layer 10 and contributed to the final prediction (Fig.~\ref{fig:bert-diagnosis}B).} 
This confirmed that \jwrevision{the spurious correlation between ``dvd'' and negative label} 
was indeed an important reason for the wrong prediction. 
\liushixia{$E_3$} thus decided to remove ``dvd'' from all 53 samples, \shixia{as class-irrelevant words can be removed in text classification without sacrificing accuracy~\cite{zhang2018learning}.} 

From the analysis of ``superficial'' and ``dvd,'' \liushixia{$E_3$} \jwR{wondered} 
\doc{whether conflicting labels} 
and \jwrevision{shortcut issues} 
might be present in other important words, accounting for most of the wrong predictions. He thus hovered \doc{over} the words in the word contribution view and found other words that have the same problems (e.g., ``attal,'' ``subconscious''). 
\jwR{All together, \zhennew{$E_3$} corrected two conflicting labels \jwrevision{(``superficial,'' ``attal'')} and removed two words \jwrevision{(``dvd,'' ``subconscious'')} with sample bias. 
After fine-tuning the model, the eight previously misclassified samples were now predicted with correct sentiments.} 

\begin{figure}[t]
\centering
\includegraphics[width=1\linewidth]{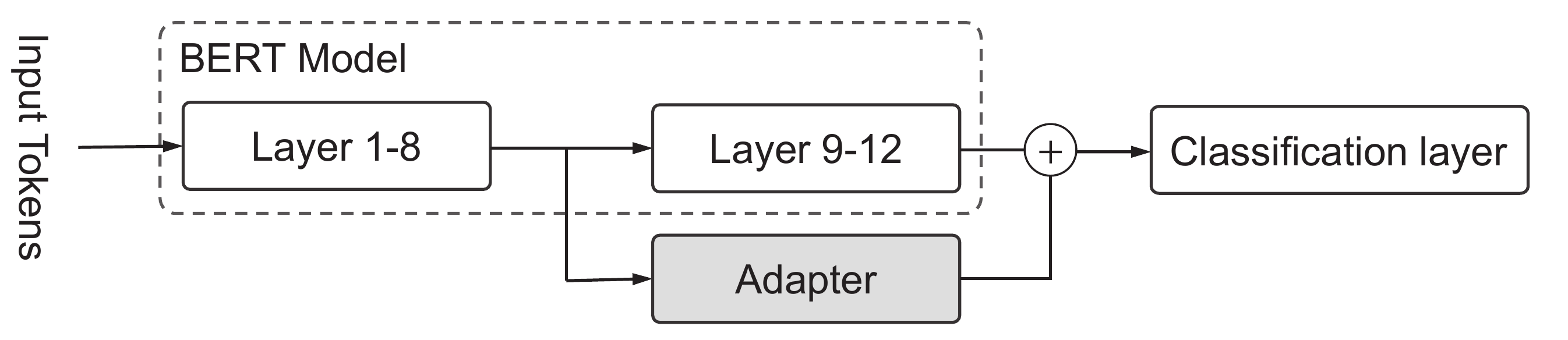}
\caption{
\zhennew{Improving the BERT model by adding an adapter module. 
}
}
\label{fig:bert-adapter}
\end{figure}

\lizhen{\zhennew{$E_3$} was also interested in other \jwR{deficiencies} 
in the model besides the data problem. He noticed that in the last \jwrevision{two} layers, 
\jwrevision{several words with strong positive or negative tendencies, such as ``love,'' ``little,'' ``bad,'' still do not contribute highly, although they have \grammarly{the} majority of the associated samples in the according sentiment class (Fig.~\ref{fig:sentiment-bert}E). For example, ``love'' occurred in 641 positive samples and 147 negative negative samples, and the similar were observed for ``little'' and ``bad.'' } 
\jwR{This indicated that the model was still coarse-grained for the specific task.}
Therefore, 
\jwR{$E_3$} adopted the adapter method to enhance the model performance by inserting an adapter module into the last four layers (Fig.~\ref{fig:bert-adapter}). 
He followed the adapter architecture proposed by Houlsby et al.~\cite{houlsby2019parameter} with the hidden layer size of 64 and the tanh activation function. The parameters of the \jwR{original} fine-tuned model were frozen, and the new adapter module was trained. 
\jwrevision{After training, the importance of ``love,'' ``little,'' and ``bad'' increased, showing that} the ability of the new model to process these \jwR{class-relevant}  
words was improved.} 
\lizhen{Accordingly, the test accuracy of the new model was increased from 93.23\% to 93.92\%.}\looseness=-1

\subsubsection{Model Comparison}

\begin{figure}[t]
\centering
 \includegraphics[width=1\linewidth]{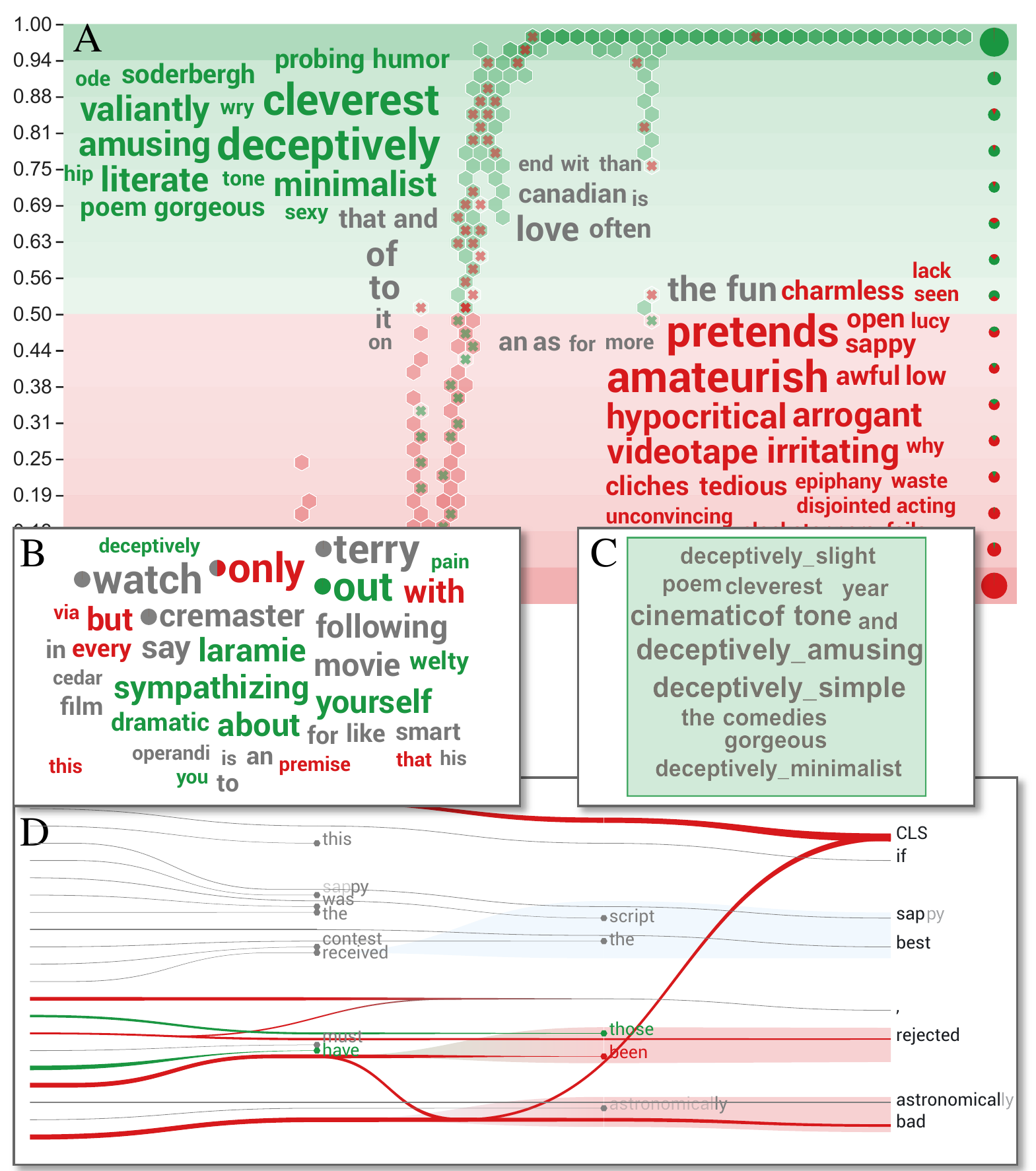}
\caption{
\zhennew{The analysis of ELMO+LSTM model. }
}
\label{fig:elmo-analysis}
\end{figure}

\liushixia{$E_4$} \doc{was} interested in using the unified measure to directly compare how different models work.
\jwrevision{He carried out the sentiment classification task on the same SST-2 dataset using two models. One is the $12$-layer BERT model. The other is the $4$-layer Bi-LSTM model with input enhanced by a pre-trained ELMo+LSTM model~\cite{peters2018deep}.} 
\jwrevision{It achieved 98.28\% accuracy on the training set and 88.53\% accuracy on the test set.} 
$E_4$ then \doc{imported} each of the models into DeepNLPVis for comparison.

\begin{figure*}[!t]
\centering
\includegraphics[width=0.9\textwidth]{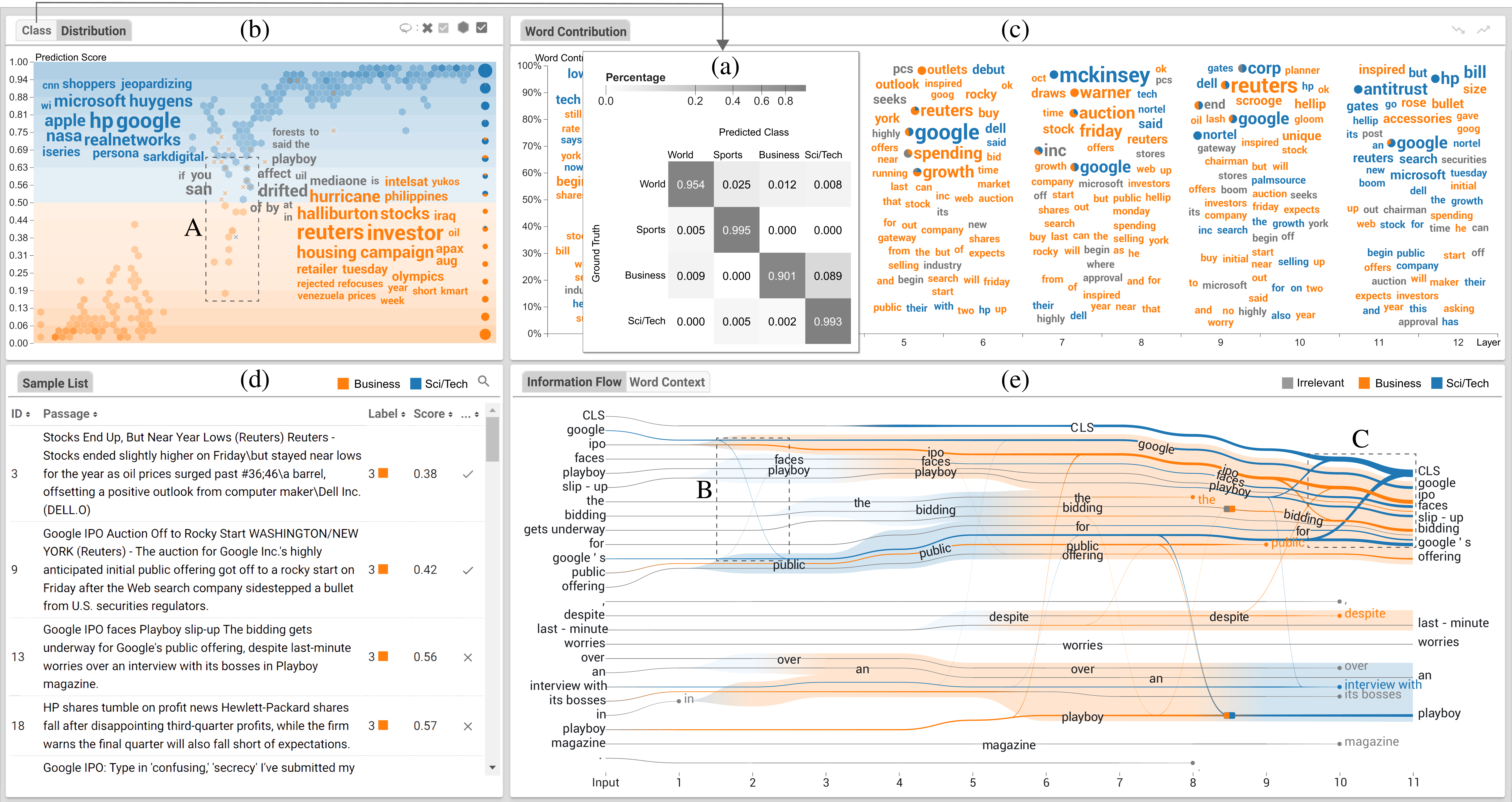}
\caption{DeepNLPVis for analyzing the BERT model on news classification: \zhennew{(a) class view for showing the overall model performance}; (b) distribution view for identifying samples and words of interest;  (c) word contribution \liushixia{of selected samples}; (d) sample list; (e) information flow for analyzing a sample by its intra- and inter-word information.}
\label{fig:teaser}
\end{figure*}

From the \lizhen{distribution} views (Fig.~\ref{fig:sentiment-bert}(a) and Fig.~\ref{fig:elmo-analysis}A), \zhennew{$E_4$} first noticed that there were many more misclassified samples using ELMo+LSTM than using BERT. Further examination showed that there was \doc{an} obvious difference in the keywords extracted.
\jwrevision{The keywords extracted in BERT are mostly common words (``laughs,'' ``recommend,'' ``boring,'' etc.), while those extracted in ELMo+LSTM included more rare words such as ``valiantly,'' ``hypocritical,'' ``amateurish.'' These words are of strong sentiments but have low frequency in the SST-2 dataset. $E_4$ considered the pretraining from ELMo helped the LSTM model to better understand the sentiments of these rare terms.}

\jwrevision{``deceptively'' was considered the most important positive keyword in ELMo+LSTM, which is against $E_4$'s intuition. Further examination of samples showed that there were far more positive samples than negative samples containing ``deceptively'' (35 positive and 1 negative). 
\jwrevision{$E_4$ thus considered ELMo+LSTM introduced the shortcut issue, the same as $E_3$ when analyzing ``dvd.''} 
To verify this, $E_4$ checked the word context for ``deceptively'' (Fig.~\ref{fig:elmo-analysis}C). Many positive words, such as ``amusing,'' ``gorgeous,'' ``cleverest,'' were presented in the context, which falsely supported the positive sentiment of ``deceptively.''
$E_4$ considered the presence of \grammarly{shortcuts} in ELMo+LSTM also explained its high accuracy on training data but a noticeable performance drop on the unseen test data.} 



\zhennew{$E_4$} then turned to analyze the misclassified samples in ELMo+LSTM. From the word contribution view (Fig.~\ref{fig:elmo-analysis}B), he found that the important words included quite a few names (e.g., ``laramie,'' \zhennew{``terry''}), nouns (e.g., ``movie,'' ``yourself'') and, stop words (e.g., \zhennew{``with,''} \jwrevision{``in''}), which are irrelevant to the prediction task. \zhennew{$E_4$} commented, ``this is consistent with the observation in \grammarly{a} previous study that the recurrent structure in LSTM limits its ability to filter some noisy words from the sequence~\cite{guan2019towards}. 
It is undesirable for sentiment classification." 
\jwrevision{$E_4$ then compared the information flow of sample \#145 using the two models. He observed that the information transition in ELMo+LSTM (Fig.~\ref{fig:elmo-analysis}D) was more local, showing the limitation of the ELMo+LSTM model in dealing with the association in non-adjacent words. 
Unlike BERT, the successful prediction was largely based on several strong negative words, such as ``reject'' and ``bad,'' rather than recognizing the turning structure of the sample. }

\jwrevision{The comparison showed superior performance of BERT over ELMo+LSTM on the sentiment classification task. $E_4$ thus considered to improve the ELMo+LSTM model using the knowledge distilled from the BERT model. He followed the knowledge distillation method in~\cite{hinton2015distilling}, where the ELMo+LSTM model was the student model while the BERT was the teacher model. The idea of knowledge distillation is to refine the student model by minimizing the difference (measured using KL divergence
) between its output and the teacher model's output, and thus to improve the performance of the student model. In this case, after knowledge distillation, the test accuracy of the new ELMo-LSTM model was increased from 88.53\% to 89.33\%.}

\subsection{\shixia{Multi-Class Classification}}

\zhennew{$E_3$} was satisfied with the assistance of DeepNLPVis for analyzing sentiment classification. To see how this assistance generalize\zhengyan{s} to other classification tasks, he \shixia{performed} a further classification task on news. A subset \doc{was} 
randomly selected from the AG News topic classification dataset~\cite{zhang2015characterlevel}. 
There are four classes: ``world'', ``sports'', ``business'', ``sci/tech'', each containing 30,000 training samples and 1,900 testing samples. 
The BERT model \doc{achieved} 94.79\% test accuracy.

To unify the analysis, a confusion matrix (Fig.~\ref{fig:teaser}(a)) \doc{was} 
provided \shixia{to select} two classes for further analysis.
\zhennew{$E_3$} observed 
that the \doc{greatest} 
confusion was between the ``business'' and the ``sci/tech'' news. 
He selected the two classes and then, following \doc{a} 
similar process in the first case, started the analysis with the \lizhen{distribution} view \jwR{(Fig.~\ref{fig:teaser}(b))}. 
\doc{He} 
observed that most samples of the two classes \doc{were} 
distributed apart (left and right). However, a small \doc{number} 
of ``business'' news (orange) \doc{were} 
horizontally closer to the ``sci/tech'' distribution (blue), indicating \jwR{these ``business'' news have some similarities to ``sci/tech'' news} 
(Fig.~\ref{fig:teaser}A). 
\doc{It seemed consequential that there were} 
more ``business'' news misclassified as ``sci/tech'' news than the other way \doc{around}. \zhennew{$E_3$} then selected these ``business'' news samples in the middle \jwR{of the distribution view} for further examination in the word contribution view (Fig.~\ref{fig:teaser}(c)). He noticed that there \shixia{was} a \doc{significant number} 
of important words that are typically related to technology, such as ``antitrust,'' ``google,'' ``hp,'' etc. 
Checking in the sample list, he further found many of these samples were ``business'' news related to ``sci/tech'' companies. 
Their content is indeed \jwR{similar} to ``sci/tech'' news.

Among these technology words, ``google'' was of high importance throughout the layers and \doc{drew} 
the attention of \zhennew{$E_3$}. 
Checking its associated samples in the sample list, there \doc{were} 
quite a few misclassified samples. Selecting the one with the lowest confidence, \zhennew{$E_3$} turned to check its information flow (Fig.~\ref{fig:teaser}(e)).
\jw{The selected sample, ``Google IPO faces Playboy slip-up The bidding gets underway for Google's public offering, despite last-minute worries over an interview with its bosses in Playboy magazine,'' \doc{was} 
a ``business'' news.}
However, ``google,'' which \doc{was} 
regarded \doc{as} a ``sci/tech'' word throughout the layers, \doc{played} an important role in the model's understanding of the sample. 
\jwrevision{The two ``googles'' had the ``sci/tech'' tendency from the beginning of the layers. When the first ``google'' transferred its information to ``google's'' at layer 2 (Fig.~\ref{fig:teaser}B), the tendency was reinforced and then passed to the [CLS] at layer 10 (Fig.~\ref{fig:teaser}C). This contributed to the misclassification.}


%% file: 6-discussion.tex
\section{Expert Feedback and Discussion}
After the case studies, \liushixia{five semi-structured interviews were conducted with the experts we worked with. 
In the interview, we first introduced the visual design and interactions, and then explored the tool together with the experts through an example case.
Each of the interview\grammarly{s} took 50-70 minutes. 
Overall, the experts gave positive feedback on the usability of DeepNLPVis.
They also pointed out several limitations that provide opportunities for future research.
}
\subsection{\liushixia{Usability}}

\liushixia{\noindent\textbf{\normalsize Informative visualization and deeper understanding}. 
All the experts agreed that the visualization was informative and helped \jwrevision{them} deeply understand the models.
They \jwrevision{particularly mentioned} 
that the polarity in the distribution view, the most contributed and \jwrevision{the} \xitingr{less contributed} words in the word contribution view, and the phrase formation in the information flow view facilitated their understanding of the models. 
$E_3$ commented, ``The polarity of contribution is very useful to detect the shortcut issue that learns spurious correlations between words and category labels (e.g., ``dvd" in the first case).
Such spurious correlations help identify which types of words (e.g., a single word, a word in context, or ordered pairs) lead to the limited diversity of training data. Knowing the \jwrevision{causes}, 
I can enhance the data accordingly."
Although \jwrevision{the experts} 
took 25.5 minutes on average (STDEV=2.89) to \jwrevision{get} familiar with the tool, \jwrevision{they} believed that the enlightening information and deeper understanding gained through the exploration deserved the efforts. 
}

\noindent\liushixia{\textbf{\normalsize Improving analysis efficiency}.
The experts especially liked the analysis process driven by a set of interactions.
They commented that existing tools, such as TensorBoard, only allowed them to examine the samples one by one.
Without a comprehensive understanding of the training process from different perspectives, diagnosing a performance issue typically relied on a time-consuming trial-and-error process.
After trying DeepNLPVis, the experts praised its efficient analysis process \jwrevision{brought by} 
the integral exploration at the corpus, sample, and word levels.
$E_1$ said, ``The analysis process 
from the sample distribution to the word contribution and the information flow inside a sample looks natural to me.
The interactions enable me to find interesting information quickly.
For example, the trending hint button in the word contribution view helps me identify the changing patterns of the words of interest, especially \jwrevision{those} conflicted with my intuition."}

\noindent\liushixia{\textbf{\normalsize Promoting effective communication in deployment}.
In the interview, the experts were \jwrevision{impressed with} 
the explanation \jwrevision{capability} 
of DeepNLPVis and the provided informative information. 
They believed that it could be used for effective communication between different sectors inside an institution. 
This is because the employed intuitive visualization provides a common ground for communication between different types of practitioners, such as model developers, consumers, and project managers.
For example, $E_4$ commented that the visual explanation provided by the information flow view could well explain how the model worked at the sample 
level.  
Such explanation is very helpful to illustrate the developed deep NLP model to the model consumers who are not machine learning experts, e.g., the developers in a product group.
With a clear understanding of the model, the developers could better maintain it in the product.
}

\subsection{\liushixia{Limitations and Future Work}}
\noindent\textbf{\normalsize \liushixia{Task Generalization}}.
In the prototype, text classification \liushixia{is used as an example} to illustrate how DeepNLPVis supports the unified understanding of NLP models. 
\liushixia{
\xitingr{Although the prototype supports the analysis of the classification tasks with a pair of sentences as input, it cannot distinguish intra-sentence and inter-sentence relations. To handle this problem, we consider designing proper visual encodings to distinguish the difference between these two relationships and enable a better analysis of such tasks.
}
In addition to classification, the experts also express the \grammarly{need} to apply DeepNLPVis to other tasks such as text \xitingr{summarization}, machine translation, and question answering. 
These tasks can be regarded as a multi-class classification task with a relatively larger class number.
In these tasks, the class number equals to the number of words/phrases.
} 
\liushixia{As a result, for these tasks, how to handle a large class number in visualization is a key challenge faced \grammarly{in the} future work.}\looseness=-1


\noindent\textbf{\normalsize Visual scalability}.
\liushixia{The experts mentioned that 
larger NLP models, such as GPT-3, usually contained dozens of layers, and a sample might contain hundreds or even thousands of words. 
With the increased \jwrevision{number of layers and sample length,} 
} the scalability issue \jwrevision{will arise in} 
the word-level \jwrevision{and sample-level} visualizations. 
A \jwrevision{possible} solution is to utilize the layer clustering technique and overview + detail visualization. 
\jwrevision{An example is the flow visualization that is affected by the sample length}\liushixia{, as each line represents a word}. 
\liushixia{This visualization} will \doc{quickly become} cluttered if hundreds or thousands of words are included in a sample. 
\liushixia{The experts indicated that \jwrevision{their analysis usually started} 
from 
representative samples and representative words in each sample.}
\jwrevision{Thus,} in the future, we are interested in identifying \jwrevision{these} representative \liushixia{samples} \jwrevision{and words} 
to balance the informativeness and readability of this \jwrevision{visualization.} 

\noindent\textbf{\normalsize Model refinement}.
After identifying the performance issues, 
\jwrevision{the experts prefer a mechanism \liushixia{that tightly integrates interactive visualization with machine learning} to refine the model semi-automatically rather than to improve the model architecture manually.} 
For example, in sentiment analysis, if the expert corrects the sentiment of several words through the visualization, s/he would expect 
the model can be \liushixia{automatically} refined based on the corrections.
Thus, 
\jwrevision{an} interesting 
\jwrevision{direction} for future work is \doc{to} explore how to transform the provided feedback into a prior or constraint \jwrevision{for} 
the model and progressively refine the model.
\xitingr{
Another related interesting direction is how to integrate active learning into the system to give more hints and reduce the number of samples to be verified by users.}

%% file: 7-conclusion.tex


\section{Conclusion}
\label{sec:conclusion}
We have presented a visual analysis method, DeepNLPVis, to facilitate a unified understanding of deep NLP models.
This method is built upon an information-based measure to illustrate how a deep NLP model maintains the information of input words in a sample with a multi-level visualization.
The effectiveness and usefulness of our method are demonstrated \doc{through} 
case studies, in which the experts utilize DeepNLPVis to understand and analyze the model behaviors in text classification tasks and explore the root \jw{causes} 
of the successful and unsuccessful cases. 
The experts are generally satisfied with the developed method as it provides a unified understanding of different deep NLP models \liushixia{allow them to conveniently compare different types of models}. 
Moreover, it helps identify the underlying reason for low performance and thus makes informed \doc{improvements in} the models.

%% file: NLPXAI-main.bbl
\begin{thebibliography}{10}
\providecommand{\url}[1]{#1}
\csname url@samestyle\endcsname
\providecommand{\newblock}{\relax}
\providecommand{\bibinfo}[2]{#2}
\providecommand{\BIBentrySTDinterwordspacing}{\spaceskip=0pt\relax}
\providecommand{\BIBentryALTinterwordstretchfactor}{4}
\providecommand{\BIBentryALTinterwordspacing}{\spaceskip=\fontdimen2\font plus
\BIBentryALTinterwordstretchfactor\fontdimen3\font minus
  \fontdimen4\font\relax}
\providecommand{\BIBforeignlanguage}[2]{{%
\expandafter\ifx\csname l@#1\endcsname\relax
\typeout{** WARNING: IEEEtran.bst: No hyphenation pattern has been}%
\typeout{** loaded for the language `#1'. Using the pattern for}%
\typeout{** the default language instead.}%
\else
\language=\csname l@#1\endcsname
\fi
#2}}
\providecommand{\BIBdecl}{\relax}
\BIBdecl

\bibitem{minaee2021deep}
S.~Minaee, N.~Kalchbrenner, E.~Cambria, N.~Nikzad, M.~Chenaghlu, and J.~Gao,
  ``Deep learning--based text classification: A comprehensive review,''
  \emph{ACM Computing Surveys}, vol.~54, no.~3, pp. 1--40, 2021.

\bibitem{kim2014convolutional}
Y.~Kim, ``Convolutional neural networks for sentence classification,'' in
  \emph{the Conference on Empirical Methods in Natural Language Processing},
  2014, pp. 1746--1751.

\bibitem{strobelt2017lstmvis}
H.~Strobelt, S.~Gehrmann, H.~Pfister, and A.~M. Rush, ``{LSTMVis}: A tool for
  visual analysis of hidden state dynamics in recurrent neural networks,''
  \emph{IEEE Transactions on Visualization and Computer Graphics}, vol.~24,
  no.~1, pp. 667--676, 2018.

\bibitem{vaswani2017attention}
A.~Vaswani, N.~Shazeer, N.~Parmar, J.~Uszkoreit, L.~Jones, A.~N. Gomez,
  {\L}.~Kaiser, and I.~Polosukhin, ``Attention is all you need,'' in
  \emph{Advances in Neural Information Processing Systems}, 2017, pp.
  5998--6008.

\bibitem{devlin2018bert}
J.~Devlin, M.-W. Chang, K.~Lee, and K.~Toutanova, ``{BERT}: Pre-training of
  deep bidirectional transformers for language understanding,'' in \emph{the
  Conference of the North American Chapter of the Association for Computational
  Linguistics}, 2019, pp. 4171--4186.

\bibitem{kim2021playing}
A.~Y. Kim and J.~Hardin, ``“playing the whole game”: A data collection and
  analysis exercise with google calendar,'' \emph{Journal of Statistics and
  Data Science Education}, vol.~29, no. sup1, pp. S51--S60, 2021.

\bibitem{yuan2021survey}
J.~Yuan, C.~Chen, W.~Yang, M.~Liu, J.~Xia, and S.~Liu, ``A survey of visual
  analytics techniques for machine learning,'' \emph{Computational Visual
  Media}, vol.~7, no.~1, pp. 3--36, 2021.

\bibitem{wang2020k}
R.~Wang, D.~Tang, N.~Duan, Z.~Wei, X.~Huang, G.~Cao, D.~Jiang, M.~Zhou
  \emph{et~al.}, ``{K-Adapter}: {I}nfusing {K}nowledge into {P}re-{T}rained
  {M}odels with {A}dapters,'' in \emph{Findings of the Association for
  Computational Linguistics: ACL-IJCNLP}, 2021, pp. 1405--1418.

\bibitem{ming2017understanding}
Y.~Ming, S.~Cao, R.~Zhang, Z.~Li, Y.~Chen, Y.~Song, and H.~Qu, ``Understanding
  hidden memories of recurrent neural networks,'' in \emph{IEEE Conference on
  Visual Analytics Science and Technology}, 2017, pp. 13--24.

\bibitem{derose2021attention}
J.~F. DeRose, J.~Wang, and M.~Berger, ``Attention flows: Analyzing and
  comparing attention mechanisms in language models,'' \emph{IEEE Transactions
  on Visualization and Computer Graphics}, vol.~27, no.~2, pp. 1160--1170,
  2021.

\bibitem{tolstikhin2021mlp}
I.~Tolstikhin, N.~Houlsby, A.~Kolesnikov, L.~Beyer, X.~Zhai, T.~Unterthiner,
  J.~Yung, D.~Keysers, J.~Uszkoreit, M.~Lucic \emph{et~al.}, ``Mlp-mixer: An
  all-mlp architecture for vision,'' \emph{arXiv preprint arXiv:2105.01601},
  2021.

\bibitem{guan2019towards}
C.~Guan, X.~Wang, Q.~Zhang, R.~Chen, D.~He, and X.~Xie, ``Towards a deep and
  unified understanding of deep neural models in {NLP},'' in
  \emph{International Conference on Machine Learning}, 2019, pp. 2454--2463.

\bibitem{dovsilovic2018explainable}
F.~K. Do{\v{s}}ilovi{\'c}, M.~Br{\v{c}}i{\'c}, and N.~Hlupi{\'c}, ``Explainable
  artificial intelligence: A survey,'' in \emph{International Convention on
  Information and Communication Technology, Electronics and Microelectronics},
  2018, pp. 0210--0215.

\bibitem{ming2019interpretable}
Y.~Ming, P.~Xu, H.~Qu, and L.~Ren, ``Interpretable and steerable sequence
  learning via prototypes,'' in \emph{ACM International Conference on Knowledge
  Discovery and Data Mining}, 2019, pp. 903--913.

\bibitem{baehrens2010explain}
D.~Baehrens, T.~Schroeter, S.~Harmeling, M.~Kawanabe, K.~Hansen, and K.-R.
  M{\"u}ller, ``How to explain individual classification decisions,''
  \emph{Journal of Machine Learning Research}, vol.~11, pp. 1803--1831, 2010.

\bibitem{ribeiro2016should}
M.~T. Ribeiro, S.~Singh, and C.~Guestrin, ````why should i trust you?''
  explaining the predictions of any classifier,'' in \emph{ACM International
  Conference on Knowledge Discovery and Data Mining}, 2016, pp. 1135--1144.

\bibitem{ren2016squares}
D.~Ren, S.~Amershi, B.~Lee, J.~Suh, and J.~D. Williams, ``Squares: Supporting
  interactive performance analysis for multiclass classifiers,'' \emph{IEEE
  Transactions on Visualization and Computer Graphics}, vol.~23, no.~1, pp.
  61--70, 2017.

\bibitem{zhang2019manifold}
J.~Zhang, Y.~Wang, P.~Molino, L.~Li, and D.~S. Ebert, ``Manifold: A
  model-agnostic framework for interpretation and diagnosis of machine learning
  models,'' \emph{IEEE Transactions on Visualization and Computer Graphics},
  vol.~25, no.~1, pp. 364--373, 2019.

\bibitem{chawla2020}
P.~Chawla, S.~Hazarika, and H.-W. Shen, ``Token-wise sentiment decomposition
  for convnet: Visualizing a sentiment classifier,'' \emph{Visual Informatics},
  vol.~4, no.~2, pp. 132--141, 2020.

\bibitem{dong2020interactive}
Z.~Dong, T.~Wu, S.~Song, and M.~Zhang, ``Interactive attention model explorer
  for natural language processing tasks with unbalanced data sizes,'' in
  \emph{IEEE Pacific Visualization Symposium}, 2020, pp. 46--50.

\bibitem{Liu2018Analyzing}
M.~Liu, J.~Shi, K.~Cao, J.~Zhu, and S.~Liu, ``Analyzing the training processes
  of deep generative models,'' \emph{IEEE Transactions on Visualization and
  Computer Graphics}, vol.~24, no.~1, pp. 77--87, 2018.

\bibitem{ji2021usevis}
X.~Ji, Y.~Tu, W.~He, J.~Wang, H.-W. Shen, and P.-Y. Yen, ``Usevis: Visual
  analytics of attention-based neural embedding in information retrieval,''
  \emph{Visual Informatics}, vol.~5, no.~2, pp. 1--12, 2021.

\bibitem{Gehrmann2020Visual}
S.~{Gehrmann}, H.~{Strobelt}, R.~{Krüger}, H.~{Pfister}, and A.~M. {Rush},
  ``Visual interaction with deep learning models through collaborative semantic
  inference,'' \emph{IEEE Transactions on Visualization and Computer Graphics},
  vol.~26, no.~1, pp. 884--894, 2020.

\bibitem{liu2019nlize}
S.~Liu, Z.~Li, T.~Li, V.~Srikumar, V.~Pascucci, and P.-T. Bremer, ``Nlize: A
  perturbation-driven visual interrogation tool for analyzing and interpreting
  natural language inference models,'' \emph{IEEE Transactions on Visualization
  and Computer Graphics}, vol.~25, no.~1, pp. 651--660, 2019.

\bibitem{karpathy2016visualizing}
A.~Karpathy, J.~Johnson, and L.~Fei-Fei, ``Visualizing and understanding
  recurrent networks,'' in \emph{the Workshop at International Conference on
  Learning Representations}, 2016.

\bibitem{li2015visualizing}
J.~Li, X.~Chen, E.~Hovy, and D.~Jurafsky, ``Visualizing and understanding
  neural models in {NLP},'' in \emph{the Conference of the North American
  Chapter of the Association for Computational Linguistics}, 2016, pp.
  681--691.

\bibitem{hohman2018visual}
F.~Hohman, M.~Kahng, R.~Pienta, and D.~H. Chau, ``Visual analytics in deep
  learning: An interrogative survey for the next frontiers,'' \emph{IEEE
  Transactions on Visualization and Computer Graphics}, vol.~25, no.~8, pp.
  2674--2693, 2019.

\bibitem{cashman2018rnnbow}
D.~Cashman, G.~Patterson, A.~Mosca, N.~Watts, S.~Robinson, and R.~Chang,
  ``{RNNbow}: Visualizing learning via backpropagation gradients in rnns,''
  \emph{IEEE Computer Graphics and Applications}, vol.~38, no.~6, pp. 39--50,
  2018.

\bibitem{strobelt2018s}
H.~Strobelt, S.~Gehrmann, M.~Behrisch, A.~Perer, H.~Pfister, and A.~M. Rush,
  ``{Seq2Seq-Vis}: A visual debugging tool for sequence-to-sequence models,''
  \emph{IEEE Transactions on Visualization and Computer Graphics}, vol.~25,
  no.~1, pp. 353--363, 2019.

\bibitem{kwon2018retainvis}
B.~C. Kwon, M.-J. Choi, J.~T. Kim, E.~Choi, Y.~B. Kim, S.~Kwon, J.~Sun, and
  J.~Choo, ``{RetainVis}: Visual analytics with interpretable and interactive
  recurrent neural networks on electronic medical records,'' \emph{IEEE
  Transactions on Visualization and Computer Graphics}, vol.~25, no.~1, pp.
  299--309, 2019.

\bibitem{tenney2020language}
I.~Tenney, J.~Wexler, J.~Bastings, T.~Bolukbasi, A.~Coenen, S.~Gehrmann,
  E.~Jiang, M.~Pushkarna, C.~Radebaugh, E.~Reif \emph{et~al.}, ``The language
  interpretability tool: Extensible, interactive visualizations and analysis
  for {NLP} models,'' in \emph{the Conference on Empirical Methods in Natural
  Language Processing}, 2020, pp. 107--118.

\bibitem{yang2020boundary}
Y.~Yang, R.~Khanna, Y.~Yu, A.~Gholami, K.~Keutzer, J.~E. Gonzalez,
  K.~Ramchandran, and M.~W. Mahoney, ``Boundary thickness and robustness in
  learning models,'' in \emph{Advances in Neural Information Processing
  Systems}, 2020, pp. 6223--6234.

\bibitem{tenney2019you}
I.~Tenney, P.~Xia, B.~Chen, A.~Wang, A.~Poliak, R.~T. McCoy, N.~Kim,
  B.~Van~Durme, S.~R. Bowman, D.~Das \emph{et~al.}, ``What do you learn from
  context? probing for sentence structure in contextualized word
  representations,'' in \emph{International Conference on Learning
  Representations}, 2018.

\bibitem{rokach2005clustering}
L.~Rokach and O.~Maimon, ``Clustering methods,'' in \emph{Data mining and
  knowledge discovery handbook}.\hskip 1em plus 0.5em minus 0.4em\relax
  Springer, 2005, pp. 321--352.

\bibitem{van2014accelerating}
L.~Van Der~Maaten, ``Accelerating t-sne using tree-based algorithms,''
  \emph{Journal of Machine Learning Research}, vol.~15, no.~1, pp. 3221--3245,
  2014.

\bibitem{maaten2008visualizing}
L.~v.~d. Maaten and G.~Hinton, ``Visualizing data using t-{SNE},''
  \emph{Journal of Machine Learning Research}, vol.~9, no.~11, pp. 2579--2605,
  2008.

\bibitem{yahav2018comments}
I.~Yahav, O.~Shehory, and D.~Schwartz, ``Comments mining with tf-idf: the
  inherent bias and its removal,'' \emph{IEEE Transactions on Knowledge and
  Data Engineering}, vol.~31, no.~3, pp. 437--450, 2018.

\bibitem{tanahashi2012design}
Y.~Tanahashi and K.-L. Ma, ``Design considerations for optimizing storyline
  visualizations,'' \emph{IEEE Transactions on Visualization and Computer
  Graphics}, vol.~18, no.~12, pp. 2679--2688, 2012.

\bibitem{liu2013storyflow}
S.~Liu, Y.~Wu, E.~Wei, M.~Liu, and Y.~Liu, ``Storyflow: Tracking the evolution
  of stories,'' \emph{IEEE Transactions on Visualization and Computer
  Graphics}, vol.~19, no.~12, pp. 2436--2445, 2013.

\bibitem{shi2010understanding}
L.~Shi, F.~Wei, S.~Liu, L.~Tan, X.~Lian, and M.~X. Zhou, ``Understanding text
  corpora with multiple facets,'' in \emph{IEEE Conference on Visual Analytics
  Science and Technology}, 2010, pp. 99--106.

\bibitem{cui2011textflow}
W.~Cui, S.~Liu, L.~Tan, C.~Shi, Y.~Song, Z.~Gao, H.~Qu, and X.~Tong,
  ``Textflow: Towards better understanding of evolving topics in text,''
  \emph{IEEE Transactions on Visualization and Computer Graphics}, vol.~17,
  no.~12, pp. 2412--2421, 2011.

\bibitem{socher2013recursive}
R.~Socher, A.~Perelygin, J.~Wu, J.~Chuang, C.~D. Manning, A.~Y. Ng, and
  C.~Potts, ``Recursive deep models for semantic compositionality over a
  sentiment treebank,'' in \emph{the Conference on Empirical Methods in Natural
  Language Processing}, 2013, pp. 1631--1642.

\bibitem{wang2019glue}
A.~Wang, A.~Singh, J.~Michael, F.~Hill, O.~Levy, and S.~R. Bowman, ``{GLUE}: A
  multi-task benchmark and analysis platform for natural language
  understanding,'' in \emph{International Conference on Learning
  Representations}, 2019.

\bibitem{brunner2019identifiability}
G.~Brunner, Y.~Liu, D.~Pascual, O.~Richter, M.~Ciaramita, and R.~Wattenhofer,
  ``On identifiability in transformers,'' in \emph{International Conference on
  Learning Representations}, 2020.

\bibitem{mccoy2019shortcut}
T.~McCoy, E.~Pavlick, and T.~Linzen, ``Right for the wrong reasons: Diagnosing
  syntactic heuristics in natural language inference,'' in \emph{the Annual
  Meeting of the Association for Computational Linguistics}, 2019, pp.
  3428--3448.

\bibitem{zhang2018learning}
T.~Zhang, M.~Huang, and L.~Zhao, ``Learning structured representation for text
  classification via reinforcement learning,'' in \emph{AAAI Conference on
  Artificial Intelligence}, 2018.

\bibitem{houlsby2019parameter}
N.~Houlsby, A.~Giurgiu, S.~Jastrzebski, B.~Morrone, Q.~De~Laroussilhe,
  A.~Gesmundo, M.~Attariyan, and S.~Gelly, ``Parameter-efficient transfer
  learning for nlp,'' in \emph{International Conference on Machine Learning},
  2019, pp. 2790--2799.

\bibitem{peters2018deep}
M.~E. Peters, M.~Neumann, M.~Iyyer, M.~Gardner, C.~Clark, K.~Lee, and
  L.~Zettlemoyer, ``Deep contextualized word representations,'' in \emph{the
  Conference of the North American Chapter of the Association for Computational
  Linguistics}, 2018, pp. 2227--2237.

\bibitem{hinton2015distilling}
G.~Hinton, O.~Vinyals, and J.~Dean, ``Distilling the knowledge in a neural
  network,'' in \emph{NIPS Deep Learning and Representation Learning Workshop},
  2015.

\bibitem{zhang2015characterlevel}
X.~Zhang, J.~Zhao, and Y.~LeCun, ``Character-level convolutional networks for
  text classification,'' in \emph{Advances in Neural Information Processing
  Systems}, 2015, pp. 649--657.

\end{thebibliography}
